\documentclass[11pt]{article}

\usepackage[preprint]{acl}

\usepackage{times}
\usepackage{latexsym}

\usepackage[T1]{fontenc}

\usepackage[utf8]{inputenc}

\usepackage{microtype}

\usepackage{inconsolata}

\usepackage{graphicx}
\usepackage{xcolor}
\usepackage{enumitem}
\usepackage{stmaryrd}
\usepackage{amsmath}
\usepackage{tikz-dependency}
\usepackage{verbatim}
\usepackage{cprotect}
\usepackage{multirow}
\usepackage[acronym]{glossaries}
\usepackage{natbib}

\usepackage{booktabs}   
\usepackage{tabularx}   
\usepackage{array}      
\newcolumntype{Y}{>{\raggedright\arraybackslash}X}
\usepackage[table]{xcolor} 

\usepackage{pgfplots}
\pgfplotsset{compat=1.18}
\usepackage{pgfplotstable}

\usepackage[most]{tcolorbox}  

\usepackage[capitalize,noabbrev]{cleveref}

\graphicspath{{../figures/}{figures/}}

\glsdisablehyper 
\newacronym{nlu}{NLU}{Natural Language Understanding}
\newacronym{nlp}{NLP}{Natural Language Processing}
\newacronym{llm}{LLM}{Large Language Model}
\newacronym{pot}{PoT}{Proof of Time}

\title{Proof of Time: A Benchmark for Evaluating Scientific Idea Judgments}

\author{
Bingyang Ye$^{1,2}$\textsuperscript{†}, 
Shan Chen$^{1,2,3}$\textsuperscript{†},  
Jingxuan Tu$^{4}$, 
Chen Liu$^{5}$, \\
\textbf{Zidi Xiong$^{1}$, Samuel Schmidgall$^{6}$, Danielle S. Bitterman$^{1,2,3}$\textsuperscript{§}} \\
\\
\textsuperscript{†}Co-first authors, \textsuperscript{§}Corresponding author: \texttt{dbitterman@bwh.harvard.edu} \\
\\
$^1$Harvard University, $^2$Mass General Brigham, $^3$Boston Children's Hospital, \\ 
$^4$Brandeis University, $^5$Yale University, $^6$Johns Hopkins University
}

\begin{document}
\maketitle

\begin{abstract}
Large language models are increasingly being used to assess and forecast research ideas, yet we lack scalable ways to evaluate the quality of models’ judgments about these scientific ideas. Towards this goal, we introduce \textbf{\gls{pot}}, a semi-verifiable benchmarking framework that links scientific idea judgments to downstream signals that become observable later (e.g., citations and shifts in researchers’ agendas). \gls{pot} freezes a pre-cutoff snapshot of evidence in an offline sandbox and asks models to forecast post-cutoff outcomes, enabling verifiable evaluation when ground truth arrives, scalable benchmarking without exhaustive expert annotation, and analysis of human–model misalignment against signals such as peer-review awards. In addition, \gls{pot} provides a controlled testbed for agent-based research judgments that evaluate scientific ideas, comparing tool-using agents to non-agent baselines under prompt ablations and budget scaling. Across 30K+ instances spanning four benchmark domains, we find that, compared with non-agent baselines, higher interaction budgets generally improve agentic performance, while the benefit of tool use is strongly task dependent. By combining time-partitioned, future-verifiable targets with an offline sandbox for tool use, \gls{pot} supports scalable evaluation of agents on future-facing scientific idea judgment tasks.
\footnote{Code, data, and evaluation scripts are at \url{https://github.com/shan23chen/proof_of_time}}
\end{abstract}

\glsresetall

\section{Introduction}
The research community invests enormous effort into producing, sharing, and debating ideas, yet the infrastructure used to evaluate scientific ideas still largely emphasizes \emph{immediate} judgments: peer review at submission time, short-horizon leaderboard comparisons, and static benchmark evaluations that are necessarily time-constrained and often idiosyncratic. These mechanisms are indispensable, but what often matters most is whether an idea \emph{stands the test of time}. Meanwhile, ``idea quality'' is hard to define and expensive to label: human judgments are costly, sometimes inconsistent, and inevitably limited by what is known at decision time. This motivates a complementary paradigm: alongside expert judgment, we ask models to make forecasts that can later be verified against naturally arriving signals. Many consequential scientific questions are intrinsically \emph{time-indexed}: which contemporaneous papers will accrue more citations, which submissions will be recognized by award committees, how benchmark metrics will evolve, or which directions a scholar will pursue next.

At the same time, ``AI for Science'' has accelerated rapidly: beyond post-hoc analysis, machine learning systems are increasingly embedded directly in discovery pipelines across domains (e.g., hypothesis generation, candidate design, and experiment steering) \citep{wang2023scientificdiscovery,karpatne2025aienabled,liu2025hypobenchsystematicprincipledbenchmarking}. High-accuracy protein structure prediction is a canonical example \citep{jumper2021alphafold}. This progress has renewed interest in \emph{metascience} \citep{munafo2017manifesto, ioannidis2015meta}, which aims to understand how science evolves, and in tools that can forecast scientific trajectories such as topic growth and emerging areas \citep{ofer2023whatnext, wang2013quantifying, chen2006citespace, kleinberg2002bursty}. As AI systems become increasingly capable of multi-step reasoning and tool use \citep{yao2023react,schick2023toolformer,wang2023agentsurvey}, a natural question emerges: can models judge which ideas will ultimate prove impactful and durable, and can we evaluate such forecasts at scale in a way that is both principled and reproducible?


We introduce \gls{pot}, a semi-verifiable benchmarking framework for evaluating models/agents' scientific idea assessment and forecasting ability through time. \gls{pot} operationalizes a simple principle: freeze the evidence available before a cutoff time, require models to forecast outcomes after the cutoff, and score predictions when the world reveals the answer. Concretely, \gls{pot} packages each instance with a pre-cutoff snapshot of evidence placed in an offline sandbox, then evaluates predictions against post-cutoff signals such as citation counts, award tiers, benchmark trajectories, and shifts in researchers' agendas. This design has three practical advantages. First, it supports \emph{verifiability}: gold labels are not subjective annotations but outcomes that can be checked later. Second, it supports \emph{scalability}: instances can be generated programmatically from public metadata and refreshed over time without requiring exhaustive expert labeling. Third, it supports \emph{misalignment analysis}: because some outcomes reflect human judgment (e.g., peer-review awards), \gls{pot} enables systematic study of where model-based assessments diverge from human processes.

\gls{pot} is also motivated by a pragmatic concern in modern scientific workflows: when and why do agentic systems help? Tool-using agents are increasingly proposed as assistants for literature review, data analysis, and scientific writing \citep{boiko2023autonomous, lala2023paperqa, mitchener2025kosmos}, yet improvements are often reported in ways that conflate task difficulty, tool access, and test-time budget \citep{schaeffer2023emergent, huang2023large}. In \gls{pot}, we study a narrower but central, cross-cutting scope: tool-using agents acting as idea judges, where tool use supports evidence exploration and aggregation in an offline sandbox. \gls{pot} provides a controlled environment to study these trade-offs because tasks can be run with and without tools, under fixed offline constraints, and with explicit limits on interaction steps. In doing so, \gls{pot} does not assume that agents are always better \citep{kim2025sciencescalingagentsystems, he2025informationtheoreticperspectiveagentic}. Instead, it quantifies how much gain agents provide, which task types benefit most from evidence exploration, and where agent overhead yields diminishing returns.

In this paper, we formalize \gls{pot} as a time-partitioned benchmark design and instantiate it across four domains spanning \textbf{impact prediction} (citations), \textbf{scientific value assessment} (peer-review awards), \textbf{research evolution} (faculty research directions), and \textbf{technological frontier forecasting} (SOTA benchmark trajectories). We evaluate a range of models under both direct-generation and tool-using agent pipelines, and we conduct controlled ablations over tool access, offline prompting, and message-budget scaling. Finally, we analyze recurrent failure modes that reveal where models struggle to translate pre-cutoff evidence into reliable forecasts and where agentic reasoning breaks down. These results position \gls{pot} as a step toward evaluating models' scientific idea judgement; that is, how well models anticipate which research directions will be most impactful in the future.

Our contributions are threefold: (1) \textbf{Benchmark design:} we propose \gls{pot}, a time-partitioned, semi-verifiable framework that links idea-level judgments to downstream signals that become observable later; (2) \textbf{Agent-native evaluation:} we introduce an offline sandbox protocol that makes tool use measurable and supports controlled ablations over tool access, structured offline prompting, and message-budget (test-time) scaling; and (3) \textbf{Empirical findings:} across 30K+ instances spanning four domains, we show that agents demonstrate large gains on tasks that require evidence exploration and aggregation, but smaller or inconsistent improvements on structured prediction tasks.

\section{Related Work}
\label{sec:related}

\gls{pot} intersects three pillars of research: (i) forecasting scientific impact and idea quality,
(ii) time-partitioned evaluation under contamination risk, and (iii) tool-using agents and
their evaluation in realistic workflows.

\subsection{Forecasting scientific impact and idea quality}
Scientometrics has long studied how papers accrue impact over time, typically modeling citations or related proxies using bibliometric features, topic signals, and network structure \citep{price1965networks, barabasi1999emergence, griffiths2004finding, wang2013quantifying, acuna2012predicting}.
More recently, work has explored whether \emph{LLMs themselves} can serve as predictive models for impact signals such as citation counts, framing citation forecasting as a text-understanding or judgment task \citep{dewinter2024chatgpt, singla2023empirical}.
At the same time, there is growing interest in the broader question of evaluating \emph{research quality} (and the limits of doing so with automated systems), including perspectives that emphasize how ``quality'' is socially mediated and difficult to reduce to a single metric \citep{thelwall2025quality, hicks2015bibliometrics}.
Relatedly, studies of LLM-assisted scholarly writing suggest that models may reproduce—and in some cases amplify—existing citation patterns and biases, raising concerns about feedback loops if model judgments become part of the scientific pipeline \citep{algaba-etal-2025-large}.
\gls{pot} differs from prior forecasting work by emphasizing \emph{benchmarking}: we provide standardized, time-partitioned tasks with frozen evidence and later-arriving labels, enabling controlled comparisons across models, agents, and budget/tool configurations.

\subsection{Temporal generalization, live evaluation, and contamination}
Web-scale pretraining makes static benchmarks increasingly vulnerable to benchmark data contamination (BDC), which can inflate scores and reduce the meaningfulness of comparisons \citep{xu2024bdcsurvey}.
A complementary line of work argues for \emph{time-aware} or continuously refreshed evaluation to preserve measurement validity as model capabilities and training corpora evolve.
Recent ``live'' benchmarks operationalize this idea by collecting new questions over time and scoring them objectively, limiting both leakage and judge bias \citep{white2024livebench, jain2024livecodebench}.
Other approaches explicitly construct contamination-resistant variants of classic test sets, including counterfactual or refreshed versions designed to reduce memorization effects \citep{zhao2024mmlucf}, and interactive evaluation frameworks that probe whether a response reflects recall versus grounded understanding \citep{yu-etal-2024-kieval}.
\gls{pot} adopts the same spirit as the ``live'' benchmark--prioritizing \emph{temporal partitioning} and frozen evidence--but targets a distinct object of evaluation: \emph{scientific research-idea assessments} tied to downstream signals (citations, awards, benchmark trajectories, and research-agenda shifts) that become observable post-cutoff.

\subsection{Agents, tool use, and evaluation in realistic workflows}
Tool-using agents are increasingly evaluated in settings where success depends on multi-step interaction, state tracking, and adherence to constraints.
Recent benchmarks emphasize realistic, time-consuming tasks and dynamic interaction loops for agents (e.g., web tasks and conversational tool use) \citep{yoran-etal-2024-assistantbench, yao2024taubench, lu-etal-2025-toolsandbox}, while surveys highlight a trend toward more realistic evaluations that measure not only success rates but also cost-efficiency, robustness, and safety \citep{yehudai2025agentevalsurvey}.
In parallel, evidence accumulates that LLM-based judging can be inconsistent or unfair across groups or styles, motivating evaluation designs that avoid subjective judges when possible \citep{wang-etal-2024-llm-not-fair}.
\gls{pot} treats \emph{agentic capability} as a central evaluation dimension: we compare direct generation against tool-using agents under controlled ablations (tool access, offline constraints, and message-budget scaling), using tasks whose labels are externally verifiable rather than judge-dependent.

\begin{figure*}[t]
  \centering
  \includegraphics[width=\textwidth]{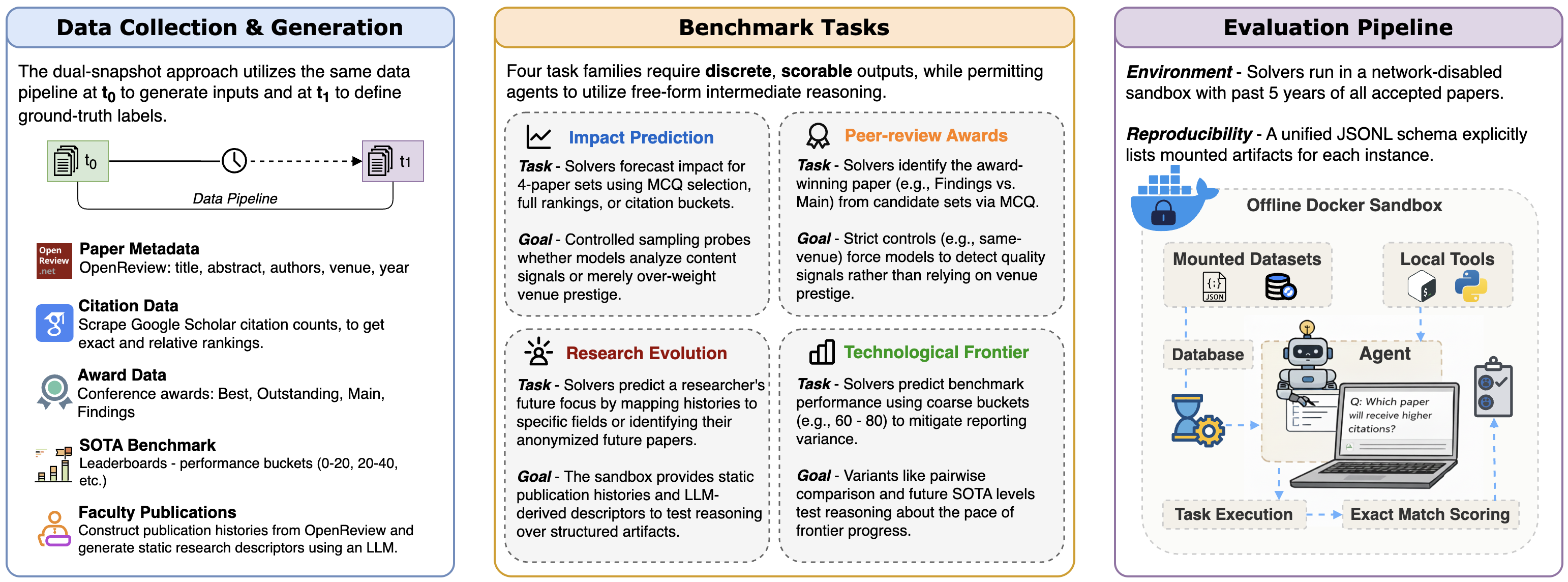}
  \caption{Overall Proof-of-Time benchmark workflow. The left panel describes our dataset creation pipeline; the middle panel emphasizes the four task families and their verifiable post-cutoff signals; the right panel illustrates offline sandbox execution with optional tool-using agents and automatic scoring.}
  \label{fig:workflow}
\end{figure*}

\section{Benchmark Design}
\label{sec:benchmark_design}

\gls{pot} is built around a simple, semi-verifiable principle:
\begin{quote}
\emph{Freeze evidence at time $t_0$; ask the model to predict signals observed at $t_1>t_0$; score once the world reveals $t_1$.}
\end{quote}
Here, ``evidence'' refers to the \emph{pre-cutoff artifacts and metadata available at time $t_0$} (e.g., paper metadata, historical counts, leaderboards) that may be informative for evaluating scientific ideas, but do not contain post-cutoff outcomes.
We call PoT \emph{semi-verifiable} because the benchmark uses verifiable downstream outcomes as imperfect proxies for idea quality. The signal is verifiable, but the target construct is not directly observable. Unlike static knowledge benchmarks, \gls{pot} treats many research-relevant judgments as inherently \emph{time-indexed}, e.g., whether an idea becomes influential, whether a paper receives an award, whether a benchmark trajectory accelerates, or whether a researcher’s agenda shifts. These outcomes are not always knowable at prediction time, but they become observable later. \gls{pot} operationalizes this by (i) constraining solvers to a pre-cutoff snapshot of evidence inside a network-isolated sandbox, and (ii) defining targets that are prospectively verifiable as time passes.

\subsection{Formal Setup: Time-Partitioned Evaluation Under Evidence Constraints}
Each \gls{pot} instance is defined by a tuple $(\mathcal{E}_{\le t_0}, q, y_{t_1})$, where $\mathcal{E}_{\le t_0}$ is the evidence snapshot available up to cutoff $t_0$, $q$ is a query over candidate items (papers, researchers, benchmarks), and $y_{t_1}$ is a label derived from a signal observed at $t_1$.
A solver receives $(\mathcal{E}_{\le t_0}, q)$ and outputs $\hat{y}$; evaluation compares $\hat{y}$ against $y_{t_1}$ once it becomes available.
Crucially, \gls{pot} separates frozen evidence, i.e., \emph{what the model may use}, from post-cutoff outcomes, i.e., \emph{what defines correctness}, thereby enabling semi-verifiable evaluation of future-facing judgments and automatic future updates.

\subsection{Offline Sandbox as a Benchmark Primitive}
To minimize leakage through opportunistic retrieval and to make tool use measurable, \gls{pot} runs solvers in a sandboxed, offline environment.
Networking is disabled, and each task instance mounts a \emph{manifested} set of read-only artifacts.
This forces agents to rely on (i) the evidence we explicitly provide and (ii) their own reasoning and analysis routines, rather than web search.
When run in agentic mode, solvers may use only a restricted tool set, including local file operations, Python analysis, and a text editor, to explore the snapshot, compute features, and justify decisions.
This design makes agent improvements interpretable: any gain must come from better use of the same frozen evidence, not from hidden access to fresher information.






\begin{table*}[t]
\centering
\small
\setlength{\tabcolsep}{8pt}
\renewcommand{\arraystretch}{1.35}
\begin{tabularx}{\textwidth}{@{}p{2.5cm} p{3.2cm} X p{2.6cm} p{2.4cm}@{}}
\toprule
\textbf{Family} &
\textbf{What it evaluates} &
\textbf{Instance setup} \newline (candidates \& evidence) &
\textbf{Outputs} &
\textbf{Verifiable signal} \\
\midrule

{\color{blue!70!black}\textbf{Impact prediction}} \newline 
\emph{(Citations)} &
Forecasting paper influence from limited pre-cutoff cues. &
\textbf{Candidates:} 4 papers from top AI conferences. \newline
\textbf{Evidence:} titles, abstracts, authors, and citation counts of historical papers (2021--2024). &
MCQ (most cited) \newline
Ranking \newline
Bucket &
Post-cutoff citation counts (Google Scholar). \\
\addlinespace[0.3em]

{\color{orange!80!black}\textbf{Peer-review Awards}} \newline 
\emph{(Awards)} &
Alignment with peer-review judgments and ability to predict award-tier outcomes. &
\textbf{Candidates:} 4 papers from top AI conferences. \newline
\textbf{Evidence:} titles, abstracts, authors, and award tiers of historical papers (2021--2024). &
Tier MCQ: Findings / Main / Outstanding / Best &
Official conference award lists. \\
\addlinespace[0.3em]

{\color{purple!70!black}\textbf{Research evolution}} \newline 
\emph{(Faculty)} &
Longitudinal reasoning about research trajectories and author/topic identification. &
\textbf{Candidates:} professor post-cutoff publication titles/abstracts with authors removed. \newline
\textbf{Evidence:} professors' pre-cutoff publications. &
Professor $\rightarrow$ field (MCQ) \newline
Paper $\rightarrow$ author (4 + None) \newline
Field focus (MCQ / open) &
Post-cutoff publications and/or curated author--field labels. \\
\addlinespace[0.3em]

{\color{green!60!black}\textbf{Technological Frontier}} \newline 
\emph{(SOTA)} &
Extrapolating benchmark progress and model performance under uncertainty. &
\textbf{Candidates:} leaderboard descriptions and metrics. \newline
\textbf{Evidence:} historical model performances. &
Bucket MCQ: $0$--$20$ / $20$--$40$ / $40$--$60$ / $60$--$80$ / $80$--$100$ &
Post-cutoff leaderboard/benchmark updates. \\

\bottomrule
\end{tabularx}
\caption{PoT task families. Each family is instantiated with automatically scorable outputs (MCQ/ranking/buckets) under an offline pre-cutoff evidence snapshot; gold labels are derived from post-cutoff, externally verifiable signals.}
\label{tab:task_families}
\end{table*}

\subsection{Four Benchmark Domains and Their Task Structure}
\gls{pot} instantiates four task families, each probing a distinct notion of future-facing research judgment and a distinct reasoning profile. In all cases, we emphasize \emph{discrete, automatically scorable outputs} (MCQ choice, ranking, or bucket), while allowing agentic solvers to use free-form intermediate reasoning internally.

\paragraph{Impact prediction.}
\textit{Citations} provides a quantitative proxy for paper influence.
We construct instances by time-partitioning the literature: the solver is given pre-cutoff evidence from papers published in top NLP venues during 2021--2024 (ACL, NAACL, EMNLP) with titles, abstracts and author information provided, together with citation counts as of October 2025, and is asked to forecast the citation counts of a \emph{candidate set} of four post-cutoff papers (titles, abstracts and authors provided) published in ACL or NAACL 2025.
\gls{pot} supports three output formats.
In \emph{MCQ} instances, the solver selects which candidate will have the most citations by the target horizon.
In \emph{Ranking} instances, the solver orders all candidates by predicted citations.
In \emph{Bucket} instances, the solver maps each paper to a coarse citation range to reduce sensitivity to minor fluctuations; we use buckets such as $0$--$10$, $10$--$50$, $50$--$200$, and $200+$.
To reduce temporal and venue confounds, each instance in \emph{MCQ} and \emph{Ranking} uses a matched candidate set of four papers drawn from the same venue--year combination, so comparisons are not driven by natural exposure effects such as earlier publication or venue-wide visibility.

\paragraph{Scientific value assessment.}
\textit{Awards} provide an externally published, human-judgment signal of research excellence and novelty.
We construct instances by time-partitioning the literature within the same three venues (ACL, NAACL, EMNLP).
The solver is given a pre-cutoff evidence pool of accepted papers from 2021--2024 with their titles, abstracts, and authors provided.
Post-cutoff evaluation papers are drawn from 2025 in the same venues, and the task is to predict the paper's award tier.
\textbf{For \textit{Awards} only}, we additionally include a diagnostic \emph{pre-cutoff} evaluation split that asks the same award-tier prediction question for papers within the evidence window, primarily to probe reliance on memorized or training-exposed information rather than evidence-grounded inference.

\paragraph{Research evolution.}
The \textit{Faculty} task family evaluates whether solvers can extrapolate from a researcher’s publication history to infer their near-future focus, capturing continuity and drift in a researcher’s agenda.
Evidence is restricted to each professor’s publications from 2021--2024, and targets are defined using 2025 publications.
We include (i) professor $\rightarrow$ field prediction (choose from a fixed taxonomy), (ii) professor $\rightarrow$ article identification (select which anonymized 2025 candidate paper belongs to the professor, with authors removed and a ``None'' option), and (iii) papers $\rightarrow$ field classification for sets of recent papers.
To support these tasks under offline constraints, the sandbox provides per-professor publication histories and auxiliary summaries such as LLM-derived keywords and area descriptors computed from pre-cutoff publications only.
This allows us to study whether agents can benefit from structured intermediate artifacts, and where such summarization introduces errors or bias.

\paragraph{Technological frontier forecasting.}
The \textit{SOTA} task measures whether solvers can reason about frontier model performance and the pace of benchmark progress.
\gls{pot} compiles normalized snapshots of popular benchmarks and leaderboards as of October 2025 and asks solvers to predict performance in coarse buckets.
We use buckets to improve robustness to prompt sensitivity and evaluation variance; for example, a five-level scale may represent performance ranges from weak to near-ceiling (e.g., $0$--$20$, $20$--$40$, $40$--$60$, $60$--$80$, $80$--$100$).
Variants include single model-benchmark bucket prediction and pairwise model comparison. We include next-year SOTA forecasting as a forward-compatible PoT task. It is not scored in this snapshot, but it becomes automatically verifiable as new leaderboard results are published.


\begin{figure*}[t]
  \centering
  \includegraphics[width=\textwidth]{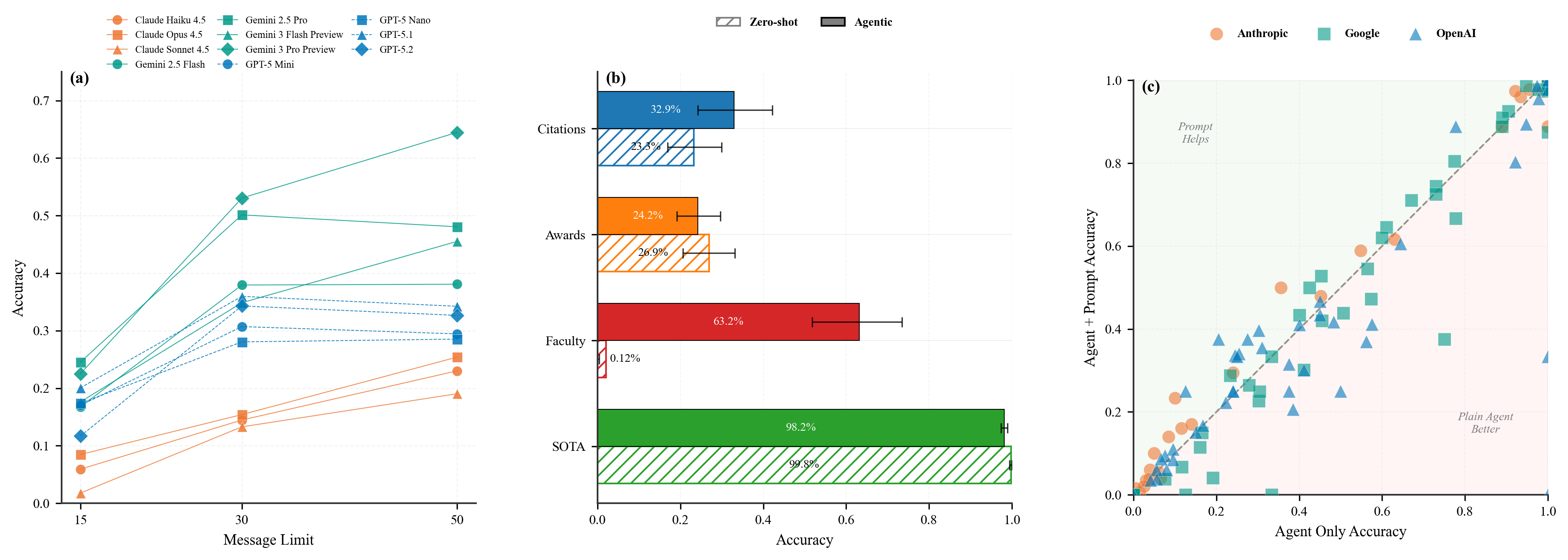}
  \caption{Core results. (A) Test-time compute scaling: accuracy vs.\ message limit (15/30/50) for each model.
(B) Task-family comparison of zero-shot vs.\ agentic performance at message limit 50, shown as average accuracy across models; error bars indicate variability across task--model combinations.
(C) Effect of adding a structured prompt on top of the same agent loop at message limit 50; points above the diagonal indicate the prompt helps.}
  \label{fig:core-results}
\end{figure*}

\section{Experiments}
\label{sec:experiments}


\subsection{Task suite and scoring}
\label{sec:exp-tasks}
We evaluate a suite of tasks spanning citations, peer-review awards, faculty research evolution, and SOTA trajectory prediction.
Tasks are designed to be automatically scorable via discrete outputs, while allowing free-form intermediate reasoning inside the sandbox.
We report exact-match accuracy for all tasks.

\subsection{Solvers: direct generation vs.\ tool-using agents}
\label{sec:exp-solvers}
We compare three solver configurations to quantify when agentic systems help,
and how much gain they provide beyond naive assumptions that agents are always better.

\begin{enumerate}
  \item \textbf{Zero-shot (direct generation).} The model answers from the prompt only,
  without tools or local data access. This tests parametric-only forecasting ability.
  \item \textbf{Agentic.} A single tool-using agent runs in a Docker sandbox with access
  to \texttt{bash}, \texttt{python}, and a text editor, iteratively gathering evidence from the offline snapshot.
  \item \textbf{Agentic + structured prompt.} Same agent loop as (2), but with an additional structured
  preamble that emphasizes offline-only operation and prescribes a more explicit tool-use protocol.
  This isolates prompt-policy effects under a fixed agent loop.
\end{enumerate}

\subsection{Test-time compute via message limits}
\label{sec:exp-messages}
To study inference-time scaling in an agentic setting, we vary a \textbf{message limit}:
the maximum number of environment interaction turns before the agent must finalize an answer.
We evaluate budgets of 15, 30, and 50 messages, corresponding to low, medium, and high test-time compute.

\begin{table}[t]
\centering
\small
\setlength{\tabcolsep}{4pt}
\renewcommand{\arraystretch}{1.1}
\begin{tabular}{>{\raggedright\arraybackslash}p{0.20\columnwidth}
                >{\raggedright\arraybackslash}p{0.72\columnwidth}}
\toprule
\textbf{Provider} & \textbf{Models} \\
\midrule
Anthropic & Claude Opus 4.5; Claude Sonnet 4.5; Claude Haiku 4.5 \\
Google    & Gemini 3 Pro Preview; Gemini 3 Flash Preview; Gemini 2.5 Pro; Gemini 2.5 Flash \\
OpenAI    & GPT-5.2; GPT-5.1; GPT-5 Mini; GPT-5 Nano \\
\bottomrule
\end{tabular}
\caption{Model suite used in our experiments. We evaluate representative frontier models across three major providers.}
\label{tab:model-suite}
\end{table}

\subsection{Models}
\label{sec:exp-models}
Table~\ref{tab:model-suite} summarizes the model suite evaluated in our experiments, covering the latest
frontier models from Anthropic, Google, and OpenAI.
All models are run under identical task definitions and scoring.
For agentic runs, the environment and tool interfaces are held constant across models.

\subsection{Implementation details}
\label{sec:exp-impl}
All agentic runs use a single-agent ReAct loop in a sandboxed environment.
We log full interaction traces, tool calls, and termination status, including runs that hit the message limit.
This enables downstream analysis of failure modes such as non-convergence and redundant exploration.

\section{Results}
\label{sec:results}

We organize results around four questions:
(i) how performance scales with test-time compute,
(ii) whether tool-using agents outperform direct generation,
(iii) whether a structured agent prompt helps beyond vanilla ReAct,
and (iv) whether conclusions change under post-cutoff (contamination-resistant) evaluation.
Unless otherwise specified, we report \textbf{sample-weighted exact-match accuracy}.

\subsection{RQ1: Test-time compute scaling}
\label{sec:rq1}
Figure~\ref{fig:core-results}A shows accuracy as a function of message limit (15/30/50) for each model in the \emph{agentic + structured prompt} setting.
Across most models, increasing the budget yields large improvements, indicating that performance is often limited
by the amount of evidence-gathering and verification the agent can perform at test time.

Scaling behavior varies across model families.
Claude models show the steepest gains, suggesting they convert additional interaction into more effective retrieval and verification.
Gemini models show strong initial performance but more moderate marginal gains.
GPT models improve with budget but tend to plateau earlier, yielding smaller gains at higher limits.
We include the full numerical scaling table in Appendix Table~\ref{tab:scaling-summary}.

\subsection{RQ2: Agentic vs.\ zero-shot performance}
\label{sec:rq2}
RQ2 asks when tool use helps under PoT's evidence constraints.
Because the frozen snapshot $\mathcal{E}_{\le t_0}$ contains only pre-cutoff context (not post-cutoff labels), any gains from agentic evaluation must come from better use of the same snapshot.

Figure~\ref{fig:core-results}B compares zero-shot and agentic performance at message limit 50, aggregated by task family.
The effect is highly task-dependent.
\textit{Faculty} shows the largest improvement, with agentic accuracy rising from near-zero in zero-shot to roughly two-thirds when tools are enabled.
\textit{Citations} shows a moderate gain (about +10pp), whereas \textit{Awards} changes little on average.
\textit{SOTA} remains near ceiling in both settings, indicating limited headroom for tool use on this coarse bucketed variant.
Overall, tool use yields its largest benefits on families where additional snapshot exploration appears most valuable, while improvements are smaller or muted when tasks are already easy or offer limited headroom.

\subsection{RQ3: Structured agent prompt effects}
\label{sec:rq3}
Figure~\ref{fig:core-results}C shows that the structured agent prompt has \emph{family-dependent} effects.
Overall, Claude models tend to benefit from the added behavioral constraints, GPT models are often neutral or slightly worse,
suggesting the prompt may be over-prescriptive, and Gemini models show mixed outcomes.
These results indicate that prompt policies are not a universal ``agent upgrade'' and that model-specific prompt tuning can matter.

\subsection{RQ4: Post-cutoff evaluation}
\label{sec:rq4}
Post-cutoff evaluation can materially change conclusions about model performance on the same capability.
Table~\ref{tab:postcutoff} summarizes shifts on \textit{Awards} from pre- to post-cutoff in the \emph{agentic + structured prompt} setting with a message limit of 50.
Importantly, pre-cutoff performance may reflect both (i) potential exposure to historically available targets and (ii) differences in difficulty or data composition between splits; post-cutoff evaluation removes the former but does not control away the latter.
The shifts are highly model-dependent: for example, \texttt{Gemini 2.5 Flash} drops (-17.7pp), whereas several GPT and Gemini variants increase by double digits.
Overall, these results show that moving evaluation beyond training cutoffs can substantially alter rankings, motivating post-cutoff splits as a standard component of Proof of Time.

\begin{table}[t]
\centering
\small
\setlength{\tabcolsep}{6pt}
\begin{tabular}{l r r r}
\toprule
Model & Pre-cutoff & Post-cutoff & $\Delta$ \\
\midrule
Claude Haiku 4.5         & 2.0\%  & 12.3\% & +10.3pp \\
Claude Opus 4.5          & 1.5\%  & 23.1\% & +21.6pp \\
Claude Sonnet 4.5        & 0.0\%  & 2.5\%  & +2.5pp \\
Gemini 2.5 flash         & 47.2\% & 29.5\% & -17.7pp \\
Gemini 2.5 pro           & 26.4\% & 33.6\% & +7.2pp \\
Gemini 3 flash           & 0.0\%  & 0.0\%  & +0.0pp \\
Gemini 3 pro             & 21.8\% & 18.8\% & -3.0pp \\
GPT-5 mini               & 8.6\%  & 32.0\% & +23.3pp \\
GPT-5 nano               & 35.5\% & 25.4\% & -10.1pp \\
GPT-5.1                  & 16.8\% & 32.0\% & +15.2pp \\
GPT-5.2                  & 3.5\%  & 28.7\% & +25.1pp \\
\bottomrule
\end{tabular}
\caption{Pre-cutoff vs.\ post-cutoff award-tier prediction accuracy in \emph{agentic + structured prompt} setting at message limit 50.}
\label{tab:postcutoff}
\end{table}

\section{What Goes Wrong in Agentic Runs?}
\label{sec:agent-behavior}

To understand why additional interaction budget does not always translate into correct forecasts, we analyze 1,759 single-agent execution traces using an LLM-as-judge protocol (Gemini 3 pro) with stratum-specific rubrics (Appendix Fig.~\ref{fig:judge-rubrics}). The judge inspects each interaction log and assigns each run to one of three strata: \emph{complete-correct}, \emph{complete-wrong}, or \emph{incomplete} (terminated by hitting the interaction budget).
Overall, 630 runs are complete-correct (35.8\%), 625 are complete-wrong (35.5\%),
and 504 are incomplete (28.7\%).

\subsection{Why runs succeed}
\label{sec:aba-success}
Based on the LLM-as-judge annotations, complete-correct runs typically follow a concise ``retrieve--verify--commit'' pattern:
the agent identifies the relevant artifact in the offline evidence (e.g., the correct row, paper identifier, or leaderboard entry),
extracts the instance-specific signal, and commits only after the instance-local evidence is in hand.
However, the LLM judge frequently notes \emph{process-level fragility} even among correct answers, including
redundant steps and occasional ``lucky'' correctness (Appendix \Cref{tab:aba-correct-flags}).
This suggests that outcome correctness alone can mask brittle reasoning paths, motivating trace-level analysis.

\subsection{Why runs fail}
\label{sec:aba-fail}
For complete-wrong runs, the dominant failure modes are \textbf{reasoning errors} (37.7\%)
and \textbf{retrieval/tooling failures} (36.3\%), followed by
\textbf{stopping too early} (8.6\%) and
\textbf{format/constraint violations} (7.5\%).
Table~\ref{tab:aba-wrong-tax} summarizes these categories.
Qualitatively, failures often arise when the agent retrieves broadly relevant materials but misattributes them to the target instance, or when it commits based on partial evidence without a final verification step.

\begin{table}[t]
\centering
\small
\setlength{\tabcolsep}{6pt}
\begin{tabularx}{\linewidth}{Y r r}
\toprule
Category & Count & Share \\
\midrule
Reasoning error & 232 & 37.7\% \\
Retrieval/tooling & 223 & 36.3\% \\
Stopping too early & 53 & 8.6\% \\
Format/constraint violation & 46 & 7.5\% \\
Task misunderstanding & 32 & 5.2\% \\
Evidence misread & 29 & 4.7\% \\
\bottomrule
\end{tabularx}
\caption{Failure taxonomy for complete-wrong runs (primary category from the judge).}
\label{tab:aba-wrong-tax}
\end{table}

\subsection{Why runs are incomplete}
\label{sec:aba-incomplete}
Incomplete runs are primarily characterized by \textbf{looping/thrashing} and \textbf{budget exhaustion}.
Table~\ref{tab:aba-nonconv} shows that looping behaviors account for the largest share of incompletions,
either directly or by consuming the budget through repeated retrieval and debugging cycles.
In many traces, the judge identifies a clear \emph{information bottleneck} (most commonly retrieval/search formulation or parsing),
but the agent fails to reach a decisive ``last-mile'' step before the message limit.

\begin{table}[t]
\centering
\small
\setlength{\tabcolsep}{6pt}
\begin{tabularx}{\linewidth}{Y r r}
\toprule
Category & Count & Share \\
\midrule
Looping / thrashing & 183 & 36.5\% \\
Budget exhaustion & 145 & 28.9\% \\
Budget exhaustion via looping & 100 & 19.9\% \\
Other & 45 & 9.0\% \\
Tooling / environment errors & 22 & 4.4\% \\
Missing resource / data & 6 & 1.2\% \\
Unresolved ambiguity & 1 & 0.2\% \\
\bottomrule
\end{tabularx}
\caption{Non-convergence categories for incomplete runs.}
\label{tab:aba-nonconv}
\end{table}

\section{When Do Agents Pay Off?}
\label{sec:when_to_use_agents}

Tool-using agents can improve PoT performance, but the gains are rarely free: in our runs, agentic execution often increases token consumption by one to two orders of magnitude relative to a zeroshot baseline, and the resulting accuracy improvements are highly model- and task-dependent. We summarize these trade-offs using efficiency frontiers that show accuracy gain over zeroshot against token overhead at multiple message budgets. The results show that at low budgets, several model configurations yield negligible or even negative gains despite substantial overhead, while higher message limits can unlock improvements for some models but not others. Aggregating by task family provides a complementary view: evidence-intensive families such as \textit{Faculty} and \textit{Citations} benefit most from increased interaction/compute budgets, whereas \textit{Awards} and \textit{SOTA} exhibit smaller or less reliable gains under comparable overhead. Overall, PoT supports a pragmatic policy: use zeroshot or low-budget agents for routine tasks, and reserve higher-budget agents for cases requiring extensive evidence gathering, critical verification, or where errors are costly. 
Detailed results  are shown in Figure~\ref{fig:agent_vs_zeroshot_model} and 
Figure~\ref{fig:agent_vs_zeroshot_family} in Appendix~\ref{app:agent_cost_efficiency}.

\section{Conclusion}
\label{sec:conclusion}
We introduced \gls{pot}, a time-partitioned, semi-verifiable benchmark framework for evaluating research-idea judgments through downstream signals that arrive after a cutoff time. By freezing pre-cutoff artifacts in an offline sandbox and scoring forecasts against post-cutoff outcomes, \gls{pot} enables scalable, contamination-aware evaluation without relying on exhaustive expert annotation. Empirically, across 30K+ instances spanning four domains, we find that increased interaction budgets generally improve agentic performance, but gains vary by model family. Tool use yields the largest improvements on \textit{Faculty} and moderate gains on \textit{Citations}, while \textit{Awards} changes little on average and \textit{SOTA} remains near ceiling in both zero-shot and agentic settings. We also observe that structured prompting is not a universal upgrade and that post-cutoff evaluation can materially shift conclusions about model performance, motivating post-cutoff splits as a standard component of Proof of Time.

\section*{Limitations}

\paragraph{Proxy targets.}
\gls{pot} evaluates scientific idea judgments against downstream signals such as citation counts, award tiers, benchmark trajectories, and shifts in publication topics. These signals are externally verifiable but imperfect proxies for the underlying construct of ``idea quality'': these proxies can be influenced by visibility, community dynamics, and measurement noise, and may under-represent important contributions that accrue impact slowly or outside mainstream venues.

\paragraph{Temporal and collection choices.}
Labels depend on the choice of cutoff $t_0$ and horizon $t_1$, and on the specific data sources and joining heuristics used to construct pre- and post-cutoff snapshots. While the benchmark is designed to be refreshable, different horizons or collection pipelines may yield different difficulty profiles and error patterns.

\paragraph{Offline sandbox realism.}
The offline sandbox intentionally removes live web access to reduce leakage and make tool use measurable. This improves experimental control but departs from how researchers and deployed assistants typically operate, where up-to-date retrieval is often available. Results should therefore be interpreted as measuring performance under a constrained evidence regime rather than full open-world assistance.

\paragraph{Agent loop and compute budget.}
Our agentic solvers use a particular single-agent interaction protocol and a finite message budget. Different agent architectures, termination policies, or tool interfaces may change absolute performance and failure modes; message-limit scaling in particular may not translate directly to other interaction designs or cost regimes.

\section*{Acknowledgments}

We thank the anonymous reviewers for helpful feedback. We also thank colleagues and collaborators who provided comments on early drafts and discussions on benchmark design and evaluation.

\bibliography{custom} 

\appendix
\onecolumn

\noindent\textbf{Appendix overview.}
This appendix provides supplementary details on dataset construction (\S\ref{app:data}), experimental setup (\S\ref{app:hyperparameter}), additional experimental results (\S\ref{app:results}), compute cost (\S\ref{app:cost}), trace-level agent behavior diagnostics (\S\ref{app:aba}), trace-level LLM-as-judge diagnostics protocol (\S\ref{app:llm-protocol}), and examples for the structured prompt in our agentic workflow (\S\ref{app:prompt}).

\section{Data Collection and Dataset Construction}
\label{app:data}

\gls{pot} uses the same data collection and parsing pipelines at two time points. We first collect a snapshot at cutoff time $t_0$ and use it to construct the evidence that solvers can access in the offline sandbox. For all tasks, we define $t_0$ as January 2025 for all models whose knowledge cutoff precedes that date, which reduces the risk of data contamination during inference. The only exception is GPT-5.2, whose knowledge cutoff is August 31, 2025, so for GPT-5.2 we set $t_0$ to September 1, 2025. We later collect a second snapshot at time $t_1$ using the same procedures and use it to define labels and evaluate predictions. This design keeps formats and join keys consistent across time, while ensuring that correctness is determined by post-cutoff signals.

\subsection{Paper metadata}
We use OpenReview as the primary structured source for paper metadata. For each paper, we parse the title, abstract, author list, venue, and year, and we retain additional fields when available such as keywords and subject areas. Because different venues expose slightly different schemas, we normalize records into a common representation and construct stable paper identifiers. The $t_0$ snapshot provides the paper pools and evidence artifacts used in the sandbox. The $t_1$ snapshot is collected in the same way and is used to finalize post-cutoff paper sets and to support evaluation joins. 

\subsection{Citations data}
For \textit{Citations}, we collect citation counts from Google Scholar. We define $t_0$ as January 2025. Pre-cutoff papers are those from ACL, NAACL, and EMNLP published before $t_0$, while post-cutoff papers are ACL 2025 and NAACL 2025 papers published after $t_0$. We run the same collection pipeline at $t_1$ and store counts with retrieval timestamps, treating the $t_1$ snapshot as the gold label source for forecasting across all included papers. We set $t_1$ to November 2025. Because EMNLP 2025 is too close to $t_1$ for citation counts to meaningfully reflect longer-term impact, we exclude EMNLP 2025 papers from the citation forecasting evaluation.

\subsection{Awards data}
We parse papers from OpenReview for ACL, NAACL, and EMNLP from 2021 to 2025, along with their metadata such as title, authors, abstract, and venue. Labels are the realized award tiers, including Findings, Main, Outstanding, and Best.
We define $t_0$ as January 2025, so papers from 2021 to 2024 constitute pre-cutoff data and are eligible to serve as sandbox evidence. We define $t_1$ as November 2025, so the post-cutoff data consist of 2025 papers from the three conferences together with their award tiers, which we use for evaluation.
For ablation studies, we additionally sample 200 papers from 2021 to 2024 across ACL, NAACL, and EMNLP. These sampled papers are excluded from the historical evidence pool to avoid leakage between evidence and evaluation splits.

\subsection{SOTA benchmark snapshots}
For SOTA forecasting, we compile snapshots from mainstream benchmarks and popular public leaderboards. We normalize each snapshot into a structured schema that records the benchmark name, the metric definition, the snapshot date, and the reported score. We collect snapshots at $t_0$ and $t_1$ using the same schema.
We define $t_1$ as October 2025, corresponding to the time this paper is written.

\subsection{Professor tasks}
For professor tasks, publication histories are assembled from OpenReview paper metadata using canonicalized author names and disambiguation heuristics, and we reuse the same paper fields as above. To represent research areas in a way that is usable for prediction and offline evidence, we generate per-professor field descriptors using an LLM applied to pre-cutoff publications only. These descriptors include keywords and an area label from a fixed taxonomy, are computed once, and are stored as static artifacts in the $t_0$ sandbox. Post-cutoff updates are collected in October 2025 in the same way and are used only to define labels for evaluation, not as solver evidence.

\subsection{Task instances and sandbox manifests}
All tasks are rendered as JSONL instances with a unified schema that includes the user-facing prompt, a discrete target label or ranking, and metadata describing the task family, cutoff time, and sampling variant. Each instance also includes a manifest listing the exact read-only artifacts mounted in the sandbox at runtime. This makes evidence access explicit, supports reproducible offline evaluation, and enables post-hoc analysis grounded in the exact inputs visible to the solver.

\subsection{Task suite statistics}
Table~\ref{tab:task-stats} summarizes the number of instances per task in our evaluation suite and indicates whether labels are derived from pre-cutoff or post-cutoff signals.

\subsection{Standardized Instance Format}
All \gls{pot} tasks are represented in a common JSONL schema to support unified evaluation and cross-task comparisons.
Each record stores the prompt, the discrete target label, the solver interaction trace (for agentic runs), and metadata such as task family, variant, and message limit.
This standardization enables consistent scoring, stratified analysis by task/variant, and post-hoc error analysis grounded in the exact evidence and tool traces available to the solver at runtime.

\begin{table*}[t]
\centering
\small
\setlength{\tabcolsep}{7pt}
\renewcommand{\arraystretch}{1.15}
\begin{tabular}{%
  >{\raggedright\arraybackslash}p{0.22\textwidth}
  >{\raggedright\arraybackslash}p{0.40\textwidth}
  >{\centering\arraybackslash}p{0.10\textwidth}
  >{\raggedright\arraybackslash}p{0.16\textwidth}
}
\toprule
\textbf{Task} & \textbf{Description} & \textbf{N Samples} & \textbf{Temporal Status} \\
\midrule
Citation MCQ &
Select the most-cited paper &
200 & Post-cutoff \\
Citation Rank &
Rank papers by predicted citations &
200 & Post-cutoff \\
Citation Bucket &
Citation bucket prediction &
200 & Post-cutoff \\
Award (Pre-cutoff) &
Award tier classification  &
197 & Pre-cutoff \\
Award (Post-cutoff) &
Award tier classification  &
122 & Post-cutoff \\
Prof.\ Field &
Professor field prediction &
73 & Post-cutoff \\
Prof.\ Article &
Professor article attribution &
76 & Post-cutoff \\
Field Focus &
Field focus classification &
9 & Post-cutoff \\
SOTA Bucket &
SOTA benchmark forecasting &
45 & Post-cutoff \\
\bottomrule
\end{tabular}
\caption{Task statistics for \gls{pot} benchmark.}
\label{tab:task-stats}
\end{table*}

\section{Additional Experiment Setup}
\label{app:hyperparameter}
All models were evaluated using their default API sampling parameters (temperature, top\_p, etc.) without task-specific tuning. This ensures results reflect out-of-the-box model performance and avoids hyperparameter optimization bias. The only modified parameters were infrastructure settings: max retries (3), timeout budgets (10 min for agent tasks, 3 min for zero-shot), and concurrent connection limits (10).

\section{Supplementary Results}
\label{app:results}

\subsection{Test-time compute scaling (details)}
\paragraph{Numerical scaling summary.}
Table~\ref{tab:scaling-summary} reports the exact sample-weighted accuracies at message limits 15, 30, and 50,
along with the corresponding 15$\rightarrow$50 gains for each model. While the main text emphasizes overall
trends via Figure~\ref{fig:core-results}A, the table makes clear that scaling behavior is heterogeneous:
models vary both in their low-budget starting point and in how much they benefit from additional interaction.

\begin{figure}[t]
  \centering
  \includegraphics[width=\linewidth]{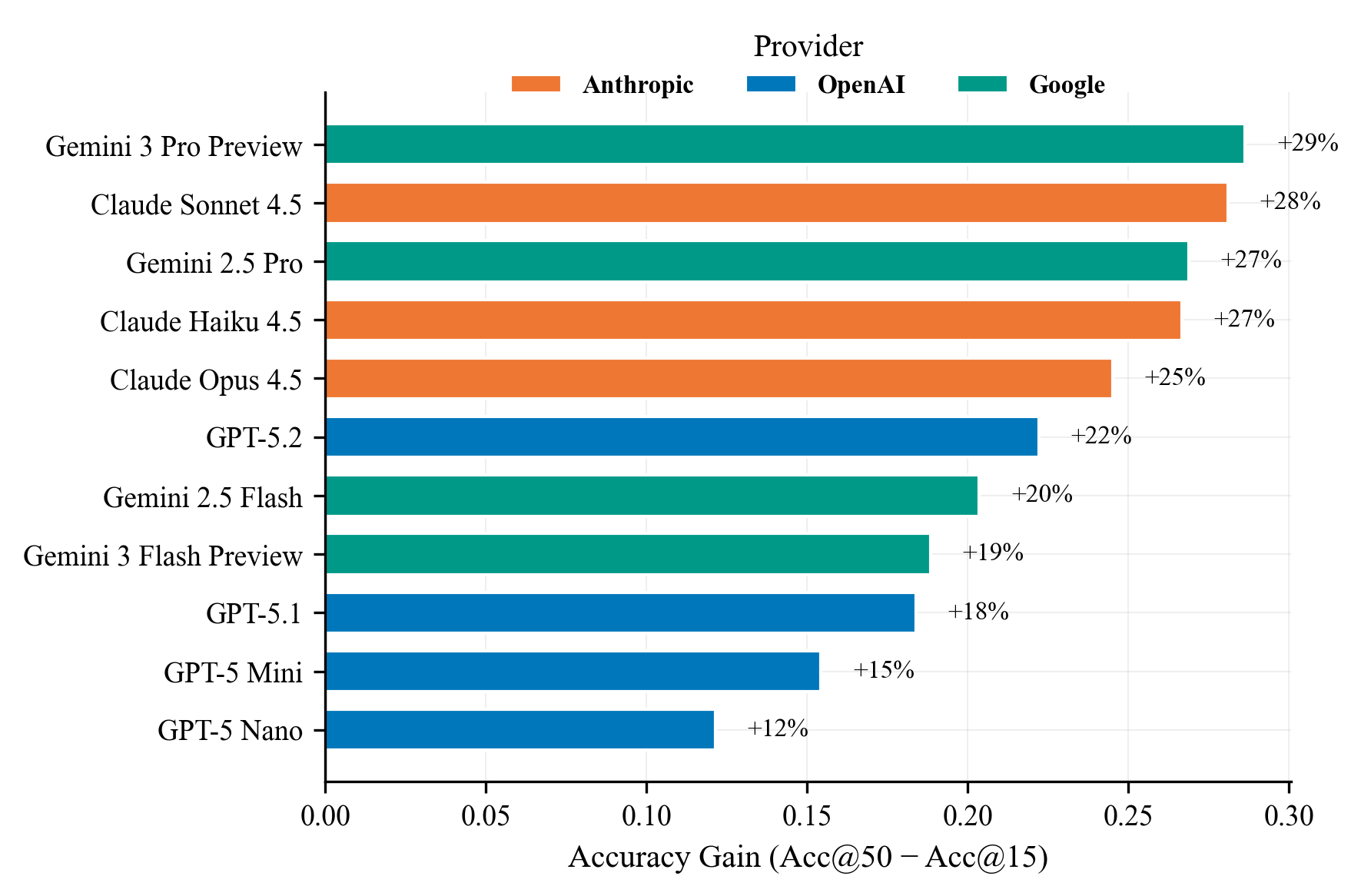}
  \caption{Scaling gains from increased test-time compute (Acc@50 -- Acc@15) by model.}
  \label{fig:scaling-waterfall}
\end{figure}

\paragraph{Gains from additional interaction budget.}
Figure~\ref{fig:scaling-waterfall} visualizes these 15$\rightarrow$50 improvements as a waterfall plot,
making it easy to see which models convert additional message budget into the largest accuracy gains.
Together with Table~\ref{tab:scaling-summary}, it shows that improvements are not uniform: some models exhibit
strong test-time scaling, while others plateau earlier and realize smaller marginal benefits.

\begin{figure}[t]
  \centering
  \includegraphics[width=\linewidth]{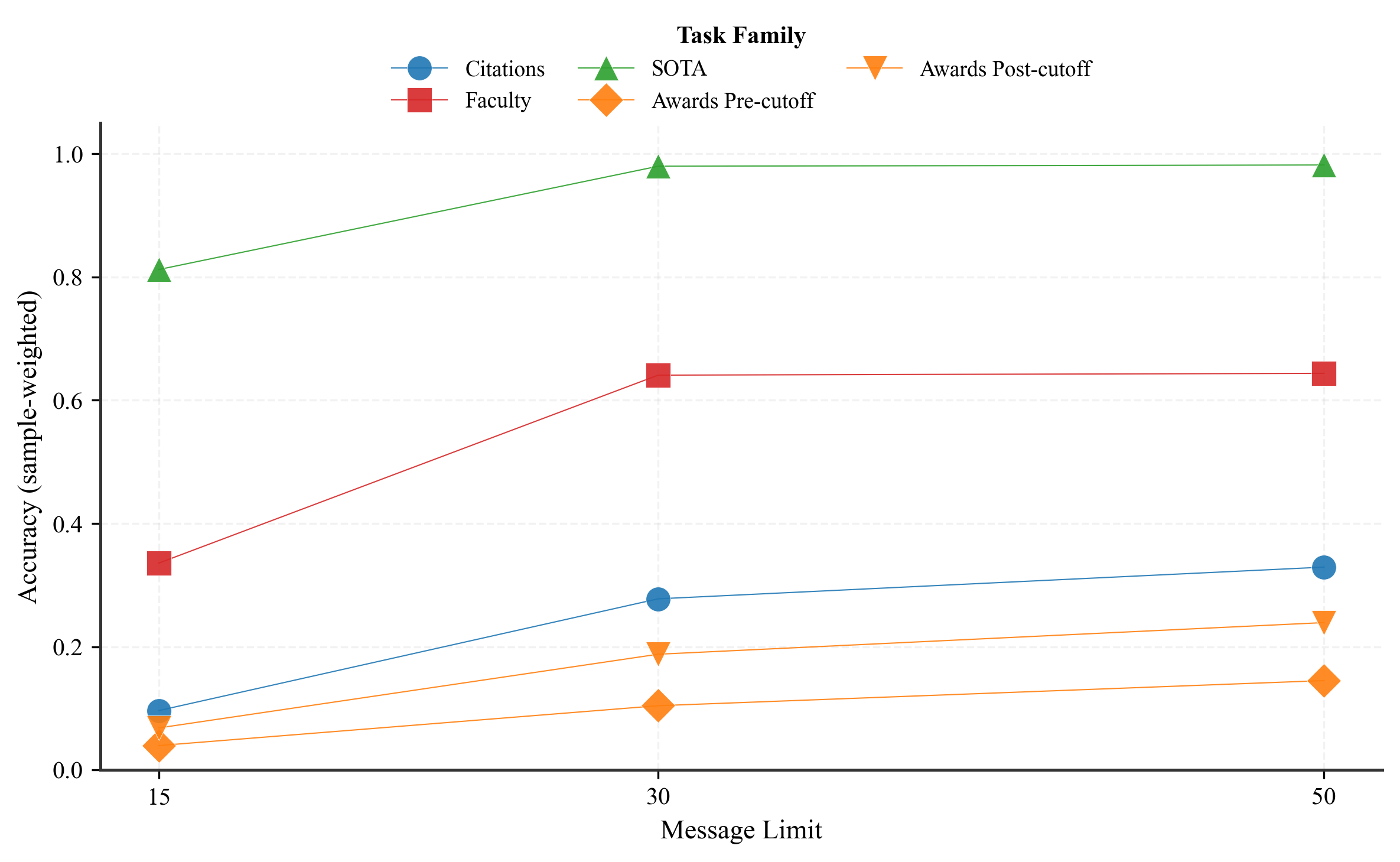}
  \caption{Scaling behavior aggregated by model family.}
  \label{fig:scaling-by-family}
\end{figure}

\paragraph{Scaling by model family.}
Figure~\ref{fig:scaling-by-family} complements Table~\ref{tab:scaling-summary} and Figure~\ref{fig:scaling-waterfall}
by aggregating the same scaling curves at the model-family level. This view smooths out model-specific variance
and highlights systematic differences in how families trade off initial accuracy versus marginal returns as the
message limit increases, providing a compact summary of provider-level scaling behavior.

\begin{table*}[t]
\centering
\small
\setlength{\tabcolsep}{6pt}
\begin{tabular*}{\textwidth}{@{\extracolsep{\fill}} l r r r r}
\toprule
Model & Acc@15 & Acc@30 & Acc@50 & $\Delta$(15$\to$50) \\
\midrule
Claude Haiku 4.5         & 8.4\%  & 26.1\% & 35.1\% & +26.7pp \\
Claude Opus 4.5          & 16.7\% & 28.2\% & 41.2\% & +24.5pp \\
Claude Sonnet 4.5        & 2.9\%  & 25.9\% & 31.0\% & +28.1pp \\
Gemini 3 Pro             & 27.2\% & 44.8\% & 55.8\% & +28.6pp \\
GPT-5.2                  & 17.9\% & 39.3\% & 40.1\% & +22.2pp \\
Gemini 2.5 pro           & 26.7\% & 55.6\% & 53.6\% & +26.9pp \\
GPT-5.1                  & 26.7\% & 44.3\% & 45.0\% & +18.4pp \\
Gemini 3 Flash           & 20.3\% & 32.0\% & 39.2\% & +18.8pp \\
Gemini 2.5 Flash         & 17.2\% & 37.0\% & 37.5\% & +20.4pp \\
GPT-5 nano               & 21.7\% & 35.1\% & 33.9\% & +12.1pp \\
GPT-5 mini               & 22.3\% & 38.7\% & 37.7\% & +15.4pp \\
\bottomrule
\end{tabular*}
\caption{Full test-time scaling summary (sample-weighted accuracy).}
\label{tab:scaling-summary}
\end{table*}

\subsection{Task interaction limits}
\begin{figure}[t]
  \centering
  \includegraphics[width=\linewidth]{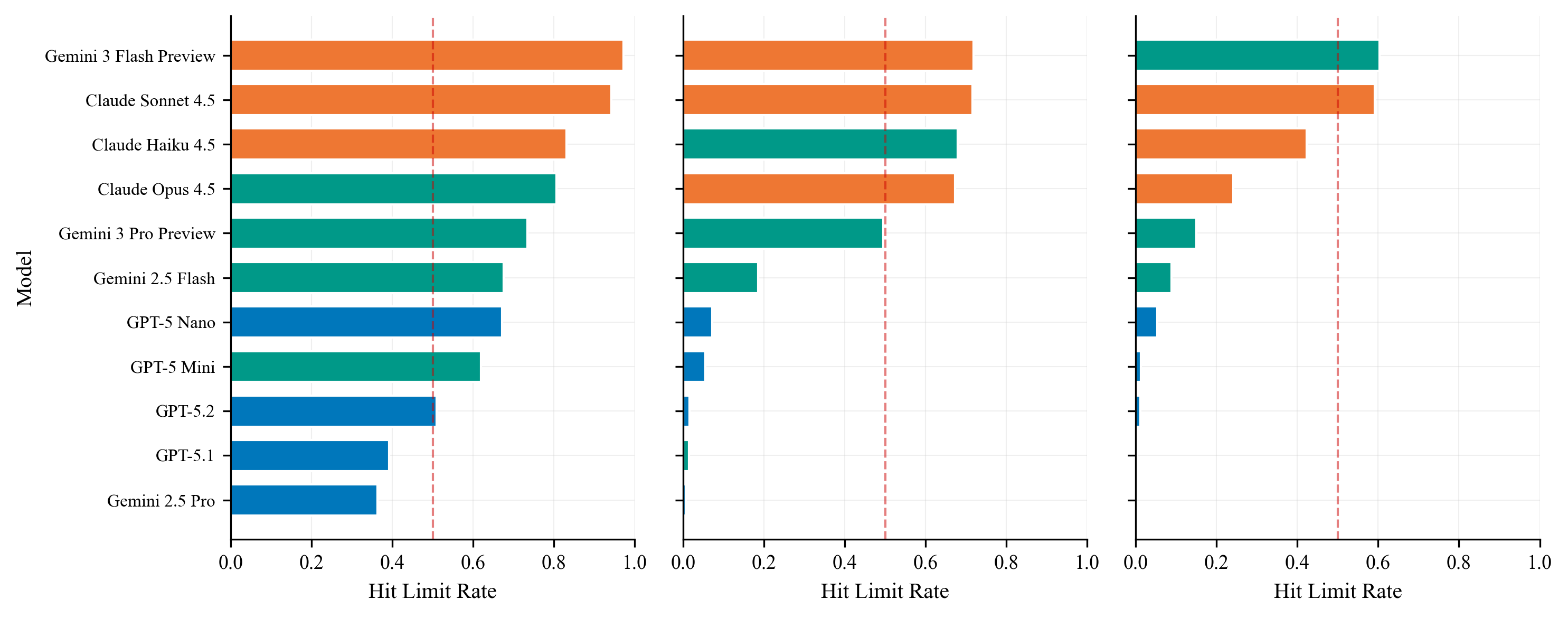}
  \caption{Analysis of runs that hit the interaction/message limit.}
  \label{fig:hit-limit}
\end{figure}

Figure~\ref{fig:hit-limit} analyzes agentic runs that terminate by exhausting the message budget rather than
by reaching a confident solution. Across models and tasks, limit hits are concentrated in the most difficult
settings, where agents repeatedly search, verify, or backtrack without converging. This highlights
non-convergence as a practical failure mode for agentic evaluation and motivates treating interaction budgets
and explicit stop rules as first-class design choices in future iterations of the benchmark.

\subsection{Agentic vs. zero-shot performance}
\begin{figure*}[t]
  \centering
  \includegraphics[width=\textwidth]{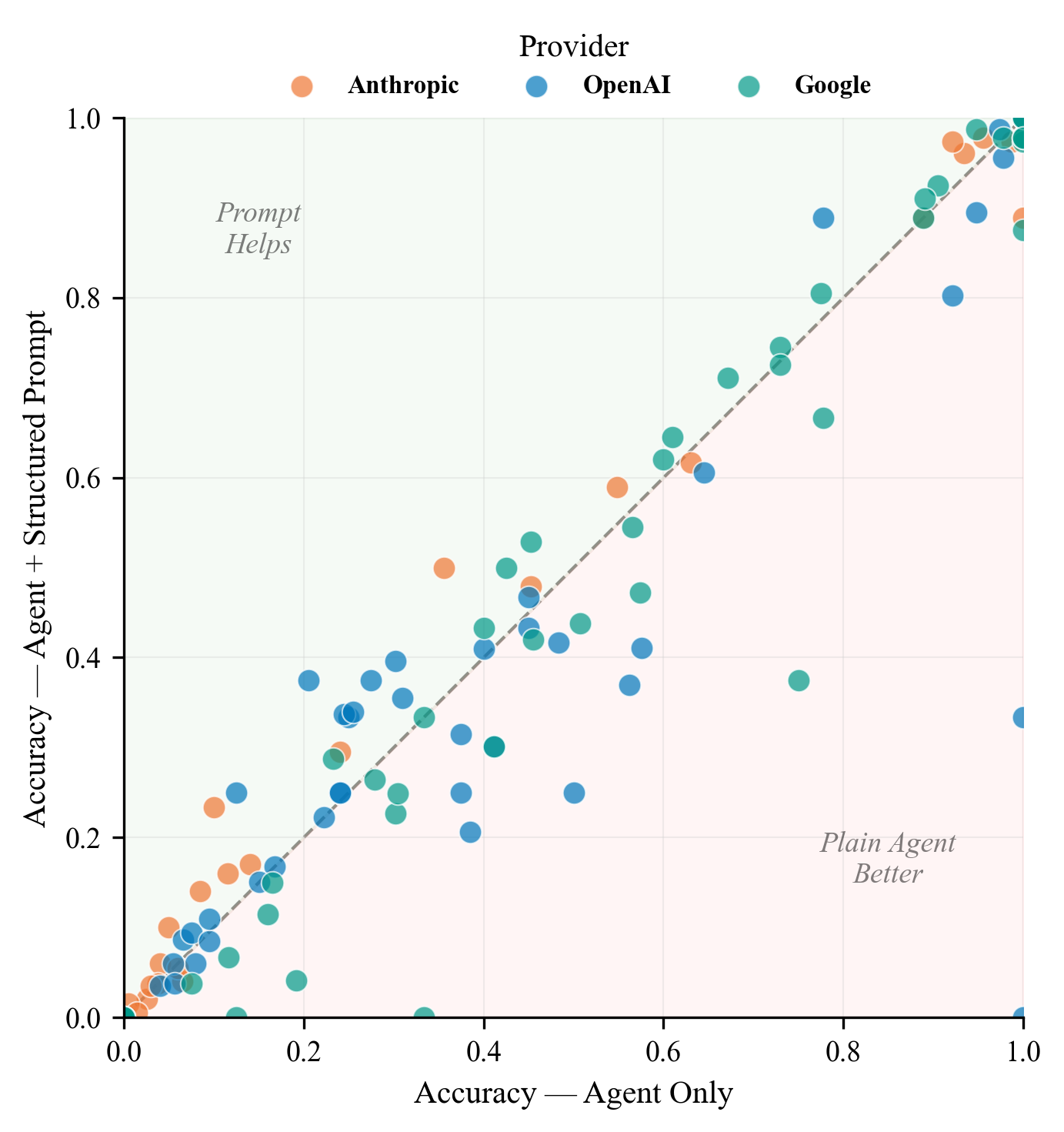}
  \caption{Agentic (message limit 50) vs.\ zero-shot accuracy across task--model combinations; points above the diagonal indicate gains from tool use.}
  \label{fig:zero_agent_scatter}
\end{figure*}

Figure~\ref{fig:zero_agent_scatter} compares agentic accuracy against zeroshot accuracy across models and tasks, with the diagonal indicating parity. Points above the diagonal correspond to settings where agentic reasoning improves performance, while points below indicate cases where direct generation is preferable. Across providers, agentic reasoning frequently outperforms zeroshot in low-baseline regimes, where zeroshot accuracy is near chance and additional interaction enables recovery from early errors. In contrast, when zeroshot accuracy is already moderate to high, gains from agentic reasoning become less consistent, and many configurations fall below the parity line, indicating that additional interaction does not reliably translate into improved decisions.

Overall, the asymmetry suggests that agentic reasoning is primarily valuable as an error-recovery mechanism rather than a uniform booster: it helps most when baseline predictions are unreliable, and offers less consistent benefits when the zero-shot system is already frequently correct.

\subsection{Model--task landscape}
\label{app:heatmap}
\begin{figure}[t]
  \centering
  \includegraphics[width=\linewidth]{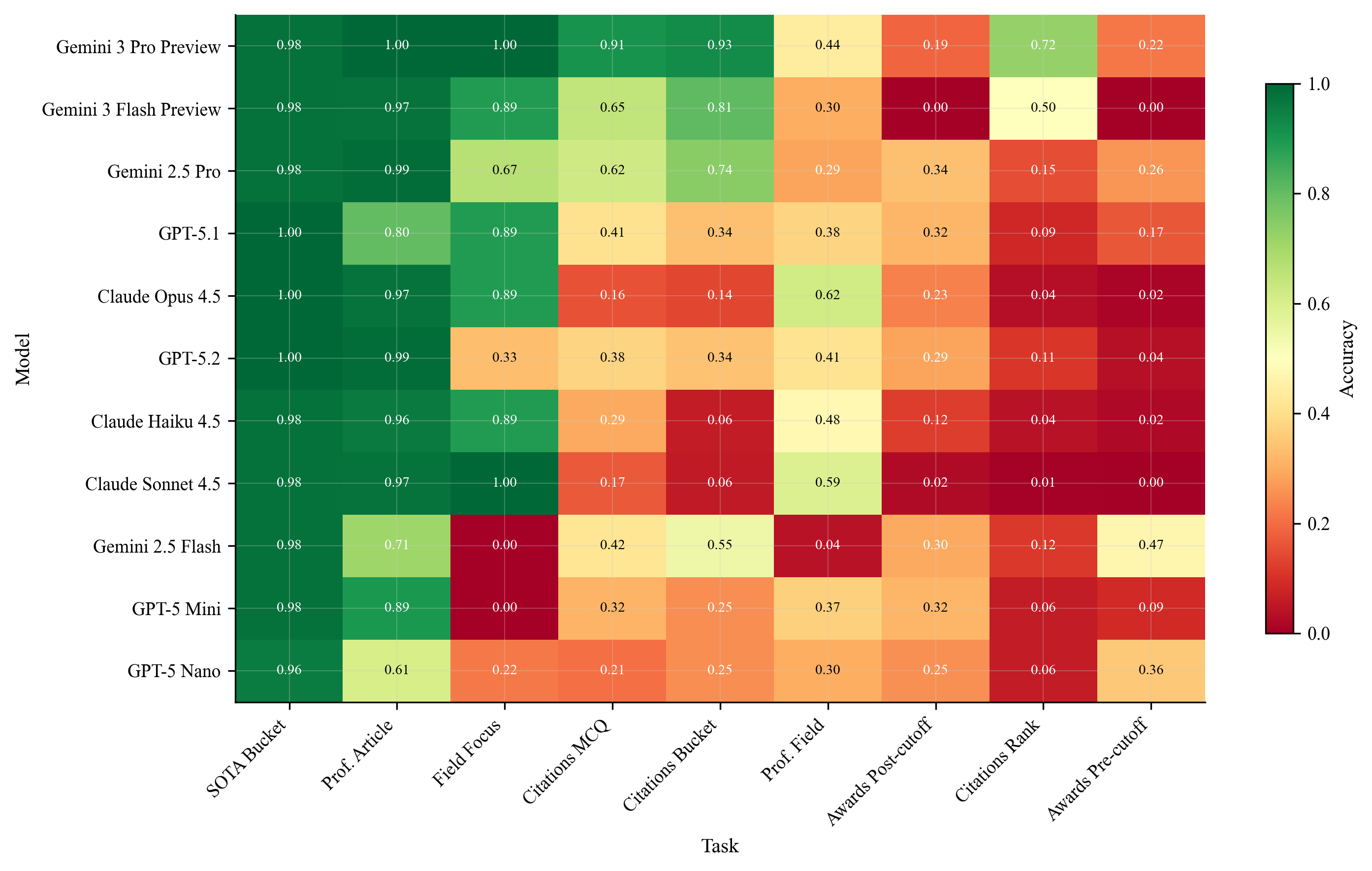}
  \caption{Model-by-task heatmap at message limit 50.}
  \label{fig:heatmap-msg50}
\end{figure}

Figure~\ref{fig:heatmap-msg50} visualizes the model--task performance landscape at message limit 50,
revealing systematic task clusters and model-family differences that are not visible from aggregate accuracy alone.

\subsection{Task difficulty}
\label{app:task-difficulty}
\begin{figure}[t]
  \centering
  \includegraphics[width=\linewidth]{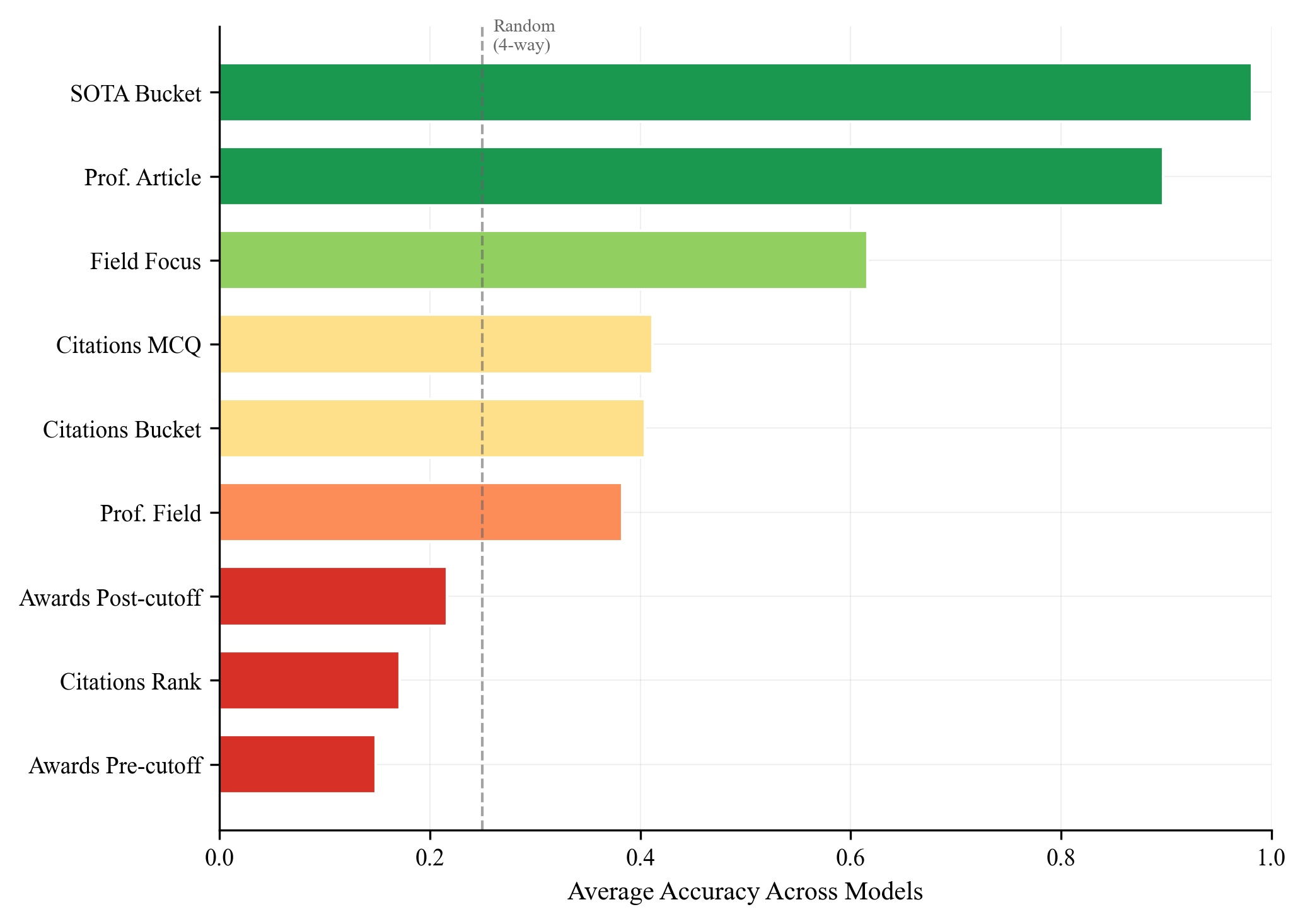}
  \caption{Task difficulty ranking averaged across models.}
  \label{fig:task-difficulty}
\end{figure}

Figure~\ref{fig:task-difficulty} summarizes task difficulty averaged across models at our primary interaction
budget. The award-tier prediction tasks are consistently among the most challenging, reflecting the noisiness
and subjectivity of peer-review outcomes as supervision signals, whereas more structured tasks such as SOTA
bucket prediction and professor--article attribution are substantially easier. This spread in difficulty
supports the benchmark design goal of covering both high-uncertainty ``idea quality'' settings and more
mechanistic, evidence-driven forecasting tasks.

\subsection{Overall performance at fixed interaction budgets}
\label{app:overall}
\begin{figure}[t]
  \centering
  \includegraphics[width=\linewidth]{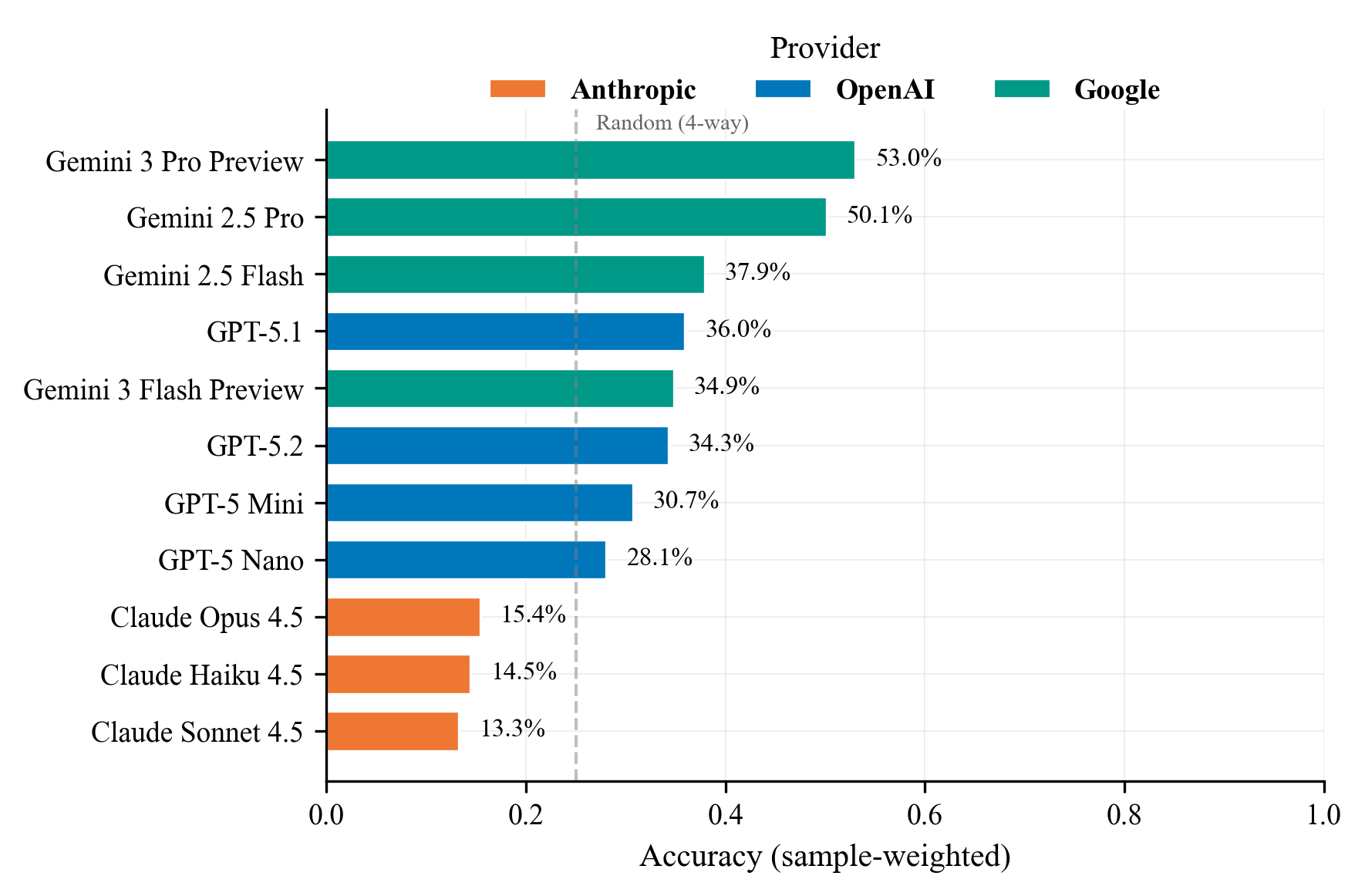}
  \caption{Overall performance at message limit 30 (mid-budget setting).}
  \label{fig:overall-msg30}
\end{figure}

\begin{figure}[t]
  \centering
  \includegraphics[width=\linewidth]{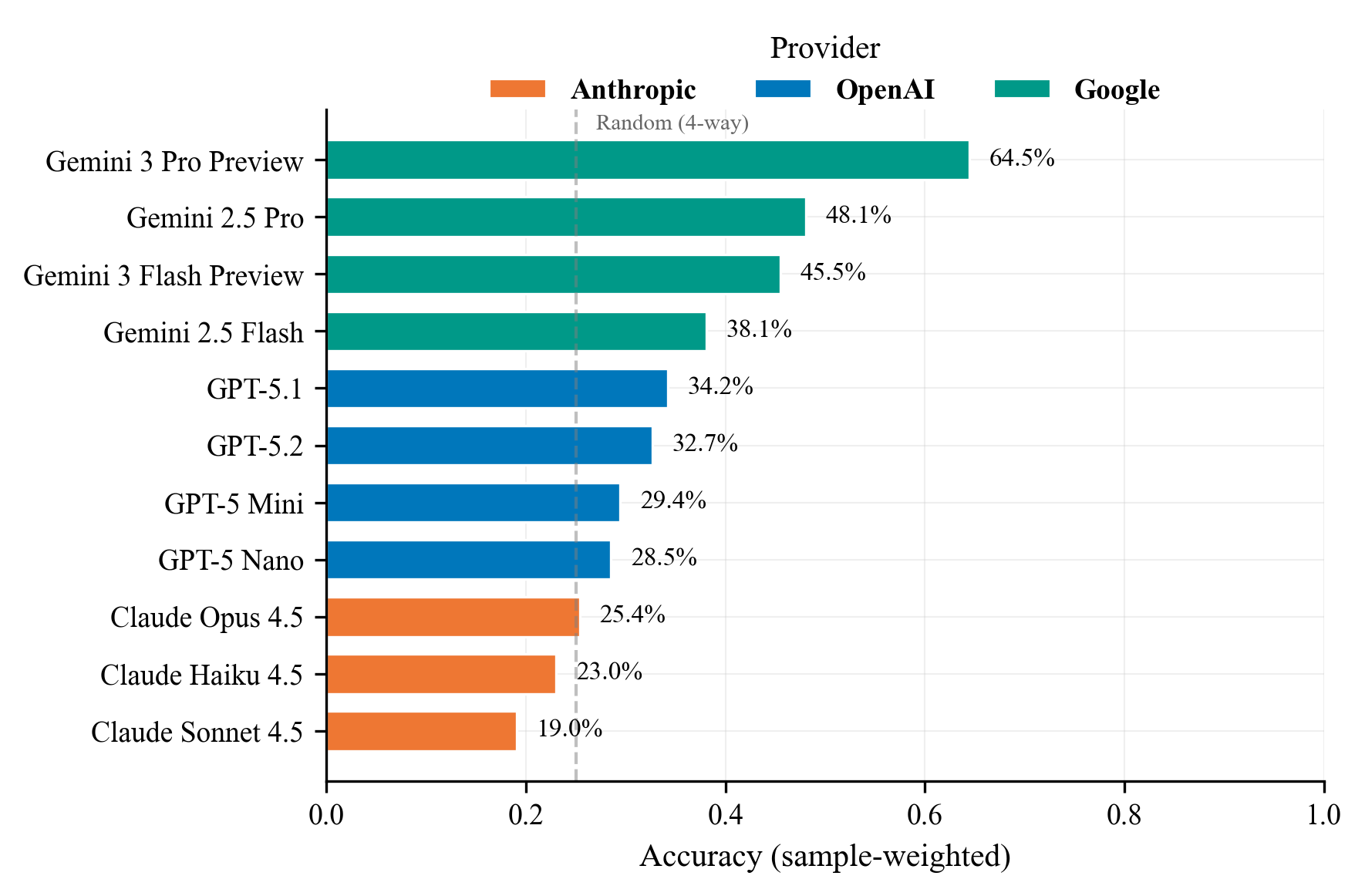}
  \caption{Overall performance at message limit 50.}
  \label{fig:overall-msg50}
\end{figure}

Figure~\ref{fig:overall-msg30} and Figure~\ref{fig:overall-msg50} provide aggregate views of model performance at message limit 30 and 50 respectively, our primary
interaction budget in the main experiments. This summary complements the scaling analysis in
Figure~\ref{fig:core-results}A by showing absolute performance at a fixed budget, and can be used as a
quick reference for cross-model comparison under a high-compute setting.

\section{Cost--Performance Analysis}
\label{app:cost}

\subsection{Compute cost}
\noindent
Table~\ref{tab:costs} reports total API spend for the runs summarized in this paper for all experiments at once. The total cost of developing our benchmark was \$11364.33. For OpenAI models we report input/output components in addition to the total; for Gemini and Claude we report totals only when token-level breakdowns were not available in our logs.

\subsection{Token usage}
To complement dollar-denominated spend, we also summarize compute consumption directly from evaluation traces as total tokens per run under different agent message budgets in Table~\ref{tab:avg_tokens_by_message_limit}. This view makes clear that agentic execution can scale sharply with interaction budget: increasing the message limit from 15 to 30 (and 50) often yields superlinear growth in tokens due to longer deliberation, tool use, and repeated context updates. In the main analysis, we therefore normalize token usage by the number of evaluated instances and include cached tokens when available, reporting “effective tokens per sample” as a model-agnostic proxy for computational overhead used in our efficiency tradeoff plots.

\begin{table}[t]
\centering
\footnotesize
\setlength{\tabcolsep}{4pt}
\renewcommand{\arraystretch}{1.15}

\begin{tabularx}{\columnwidth}{@{}l Y r r r@{}}
\toprule
\textbf{Prov.} & \textbf{Model} & \textbf{In (\$)} & \textbf{Out (\$)} & \textbf{Tot (\$)} \\
\midrule
OpenAI & GPT-5.2 & 127.565 & 84.540 & 212.105 \\
OpenAI & GPT-5.1 & 71.552  & 49.064 & 120.616 \\
OpenAI & GP-5 mini & 21.730 & 24.005 & 45.735 \\
OpenAI & GPT-5 nano & 2.578 & 11.862 & 14.440 \\
\midrule
Google & Gemini 2.5 flash & -- & -- & 91.000 \\
Google & Gemini 2.5 pro   & -- & -- & 374.000 \\
Google & Gemini 3 flash   & -- & -- & 184.000 \\
Google & Gemini 3 pro     & -- & -- & 696.000 \\
\midrule
Anthropic & Claude Haiku 4.5  & -- & -- & 395.710 \\
Anthropic & Claude Sonnet 4.5 & -- & -- & 899.540 \\
Anthropic & Claude Opus 4.5   & -- & -- & 1781.510 \\
\bottomrule
\end{tabularx}

\caption{API spend by model (USD). OpenAI totals sum input+output charges; Gemini/Claude entries report totals only.}
\label{tab:costs}
\end{table}

\begin{table}[t]
\centering
\footnotesize
\setlength{\tabcolsep}{4pt}
\renewcommand{\arraystretch}{1.15}

\begin{tabular}{l r r r r r}
\toprule
\textbf{Model} & 
\textbf{15} & 
\textbf{30} & 
\textbf{50} & 
\textbf{15$\rightarrow$30 (\%)} & 
\textbf{30$\rightarrow$50 (\%)} \\
\midrule
Claude Haiku 4.5   & 2{,}311{,}566 & 6{,}599{,}872 & 12{,}882{,}071 & 185.52 & 95.19 \\
Claude Opus 4.5    & 2{,}565{,}028 & 8{,}179{,}188 & 14{,}951{,}570 & 218.87 & 82.80 \\
Claude Sonnet 4.5  & 2{,}276{,}708 & 6{,}098{,}161 & 13{,}589{,}968 & 167.85 & 122.85 \\
Gemini 3 Pro                  & 2{,}612{,}233 & 5{,}215{,}060 & 8{,}923{,}587  & 99.64  & 71.11 \\
Gemini 3 Flash                & 1{,}060{,}000 & 2{,}050{,}000 & 3{,}450{,}000  & 93.40  & 68.29 \\
Gemini 2.5 Flash               & 1{,}973{,}369 & 2{,}881{,}910 & 3{,}373{,}690  & 46.04  & 17.06 \\
Gemini 2.5 Pro                 & 2{,}124{,}477 & 3{,}129{,}787 & 3{,}188{,}135  & 47.32  & 1.86 \\
GPT-5 mini                 & 1{,}585{,}242 & 1{,}992{,}894 & 2{,}545{,}470  & 25.72  & 27.73 \\
GPT-5 nano                 & 1{,}298{,}123 & 1{,}629{,}868 & 1{,}728{,}580  & 25.56  & 6.06 \\
GPT-5.1                    & 1{,}295{,}522 & 1{,}146{,}762 & 1{,}552{,}897  & -11.48 & 35.42 \\
GPT-5.2                    & 1{,}514{,}167 & 2{,}074{,}460 & 2{,}048{,}062  & 37.00  & -1.27 \\
\bottomrule
\end{tabular}

\caption{Average total token usage per run as a function of agent message limit.
Percentage columns report relative increases between adjacent budgets.}
\label{tab:avg_tokens_by_message_limit}
\end{table}

\subsection{Agentic cost--performance efficiency}
\label{app:agent_cost_efficiency}

\begin{figure}[t]
  \centering
  \includegraphics[width=\columnwidth]{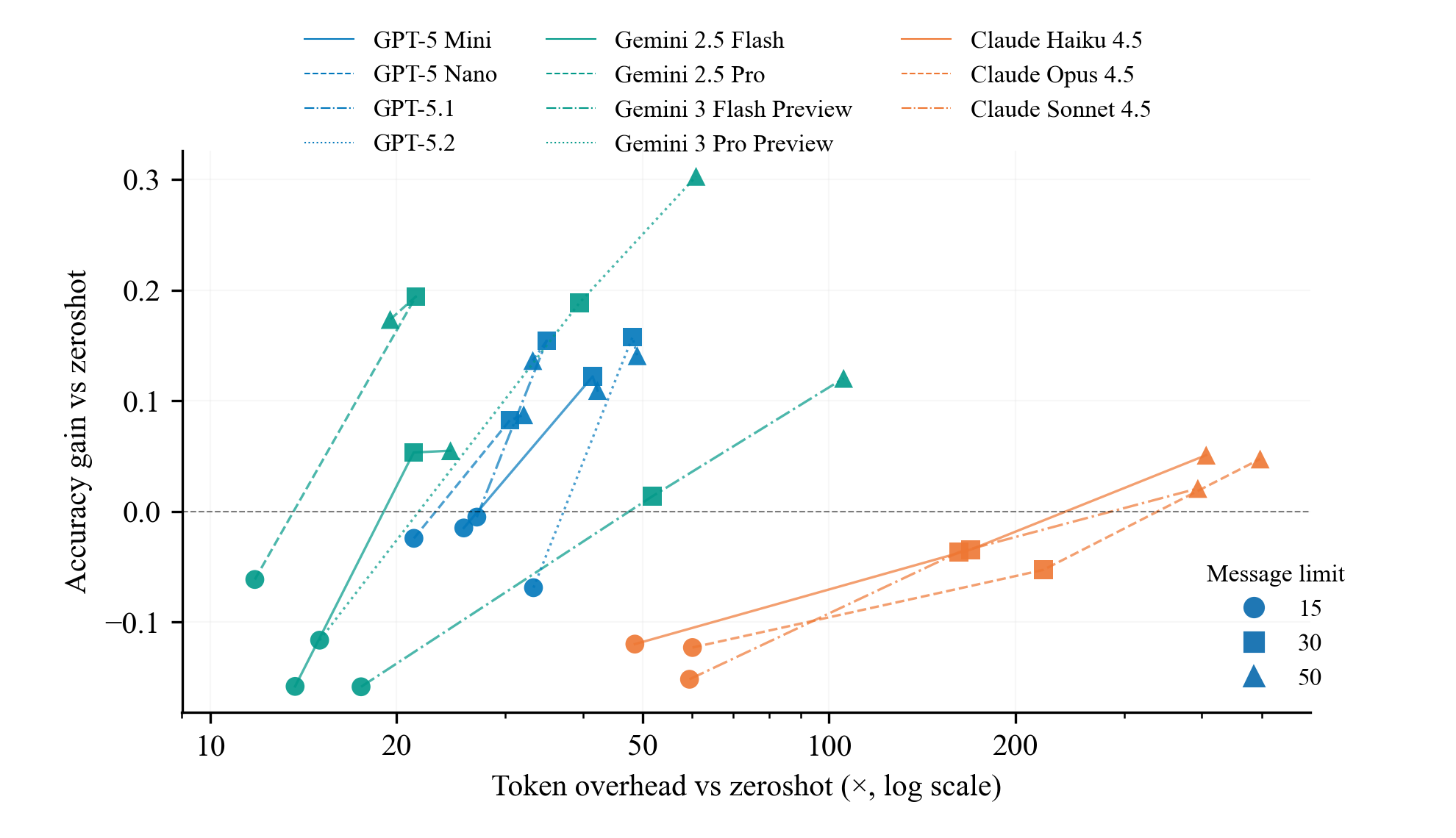}
  \caption{Model-level efficiency frontier. Each point compares an agentic configuration (15/30/50 message limit) to a zeroshot baseline for the same model, plotting accuracy gain vs.\ token overhead (log scale).}
  \label{fig:agent_vs_zeroshot_model}
\end{figure}

\begin{figure}[t]
  \centering
  \includegraphics[width=\columnwidth]{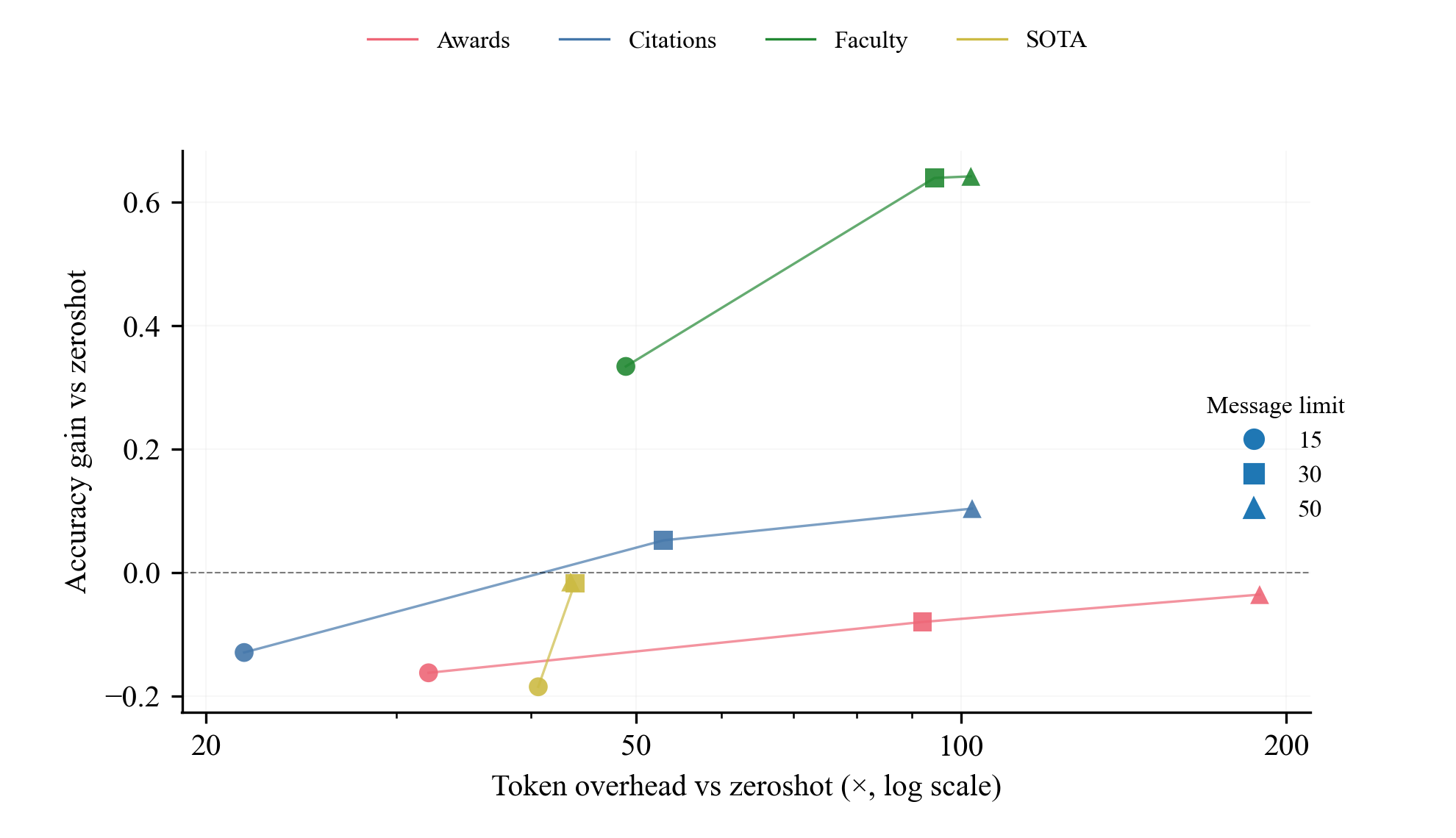}
  \caption{Task-family efficiency frontier (Awards, Citations, Faculty, SOTA). We aggregate sample-weighted accuracy within each family and compare agentic budgets against zeroshot.}
  \label{fig:agent_vs_zeroshot_family}
\end{figure}

Figure~\ref{fig:agent_vs_zeroshot_model} reports model-level efficiency frontiers, plotting accuracy gain over zeroshot against token overhead (log scale) for three interaction budgets (15/30/50 messages). Figure~\ref{fig:agent_vs_zeroshot_family} aggregates the same analysis by task family (Awards, Citations, Faculty, SOTA), highlighting that evidence-intensive families tend to benefit more from higher agent budgets.

\section{Additional Agent Behavior Diagnostics}
\label{app:aba}



\paragraph{Bottleneck categories across failures.}
Table~\ref{tab:aba-bottlenecks} aggregates the bottleneck descriptions flagged by the judge across
complete-wrong and incomplete runs. The largest category is retrieval/search, followed by parsing/extraction.
This suggests that failures are often driven by the agent's ability to (i) form effective queries and
(ii) correctly extract the instance-specific signal from retrieved artifacts, rather than by a lack of
high-level task understanding alone.


\paragraph{Process-level notes within correct traces.}
Table~\ref{tab:aba-correct-flags} summarizes heuristic flags derived from judge narratives within complete-correct runs.
Even when the final answer is correct, the judge frequently notes redundant steps and occasional lucky correctness,
highlighting that interaction traces provide a complementary signal to outcome accuracy for evaluating agentic systems.


\section{LLM-as-judge protocol for trace annotation}
\label{app:llm-protocol}
Figure~\ref{fig:judge-rubrics} lists the stratum-specific rubrics and required JSON keys used to annotate traces in the LLM-as-judge agent behavior analysis.

\begin{table}[t]
\centering
\small
\setlength{\tabcolsep}{6pt}
\begin{tabular}{l r r}
\toprule
\textbf{Bottleneck category} & \textbf{Count} & \textbf{Share} \\
\midrule
Retrieval / search                & 594 & 52.6\% \\
Parsing / extraction              & 260 & 23.0\% \\
Other                             & 136 & 12.0\% \\
Missing file / path               & 51  & 4.5\% \\
Reasoning                         & 35  & 3.1\% \\
Task understanding                & 27  & 2.4\% \\
Environment / dependencies        & 18  & 1.6\% \\
Entity disambiguation             & 8   & 0.7\% \\
\bottomrule
\end{tabular}
\caption{Information bottleneck categories flagged by the judge across failed runs (complete-wrong and incomplete).}
\label{tab:aba-bottlenecks}
\end{table}

\begin{table*}[t]
\centering
\small
\setlength{\tabcolsep}{6pt}
\begin{tabular}{l r}
\toprule
\textbf{Judge flag} & \textbf{Fraction} \\
\midrule
Verification noted          & 15.7\% \\
Unsupported claims noted    & 27.3\% \\
Lucky/guessing noted        & 43.1\% \\
Redundancy noted            & 78.6\% \\
\bottomrule
\end{tabular}
\caption{Heuristic judge flags among \emph{complete-correct} runs. Fractions are computed over correct runs only.}
\label{tab:aba-correct-flags}
\end{table*}

\begin{figure*}[t]
\centering
\setlength{\fboxsep}{6pt}
\setlength{\fboxrule}{0.4pt}

\begin{minipage}{0.3\textwidth}
\fbox{\begin{minipage}[t][5cm][t]{\textwidth}
\scriptsize\ttfamily
\textbf{Complete-correct rubric}\\
1) Grounding audit: supported vs unsupported claims.\\
2) Reasoning robustness: weakest step / luck.\\
3) Evidence sufficiency: enough evidence or partial.\\
4) Efficiency: redundant steps; minimal steps to succeed.\\
5) Generalizability: likely to work on similar instances?\\
\\
\textbf{JSON keys}\\
\{grounding\_audit, reasoning\_robustness,\\
evidence\_sufficiency, efficiency\_redundancy,\\
generalizability\}
\end{minipage}}
\end{minipage}\hfill
\begin{minipage}[t]{0.3\textwidth}
\fbox{\begin{minipage}{\textwidth}
\scriptsize\ttfamily
\textbf{Complete-wrong rubric}\\
1) First error localization: where it first goes wrong + why.\\
2) Failure taxonomy: dominant failure mode + justification.\\
3) Missing/ignored evidence: what it should have used.\\
4) Overconfidence vs uncertainty: where it should have hedged.\\
5) Minimal fix: smallest change likely to flip outcome.\\
\\
\textbf{JSON keys}\\
\{first\_error\_localization, failure\_taxonomy,\\
critical\_missing\_ignored\_evidence,\\
overconfidence\_vs\_uncertainty, minimal\_fix\}
\end{minipage}}
\end{minipage}\hfill
\begin{minipage}[t]{0.3\textwidth}
\fbox{\begin{minipage}{\textwidth}
\scriptsize\ttfamily
\textbf{Incomplete rubric}\\
1) Non-convergence cause: looping / ambiguity / over-exploration.\\
2) Last real progress: last meaningful step + what changed.\\
3) Information bottleneck: missing piece blocking completion.\\
4) Earlier commitment: where it could have answered earlier.\\
5) Progress signal: measurable progress or stagnation?\\
\\
\textbf{JSON keys}\\
\{non\_convergence\_cause, last\_real\_progress,\\
information\_bottleneck, earlier\_commitment\_point,\\
progress\_signal\}
\end{minipage}}
\end{minipage}

\caption{Stratum-specific LLM-as-judge rubrics and required JSON output keys used for trace annotation.}
\label{fig:judge-rubrics}
\end{figure*}

\section{Agent Prompt}
\label{app:prompt}

\subsection{Structured Prompt}
The structured system prompt used for the \textit{agent + structured prompt} setting is inspired by the prompt design of Google’s Antigravity agent. The original Antigravity prompt specifies a general-purpose, agentic assistant designed to support research and engineering workflows by combining structured instructions, tool use, and explicit behavioral constraints.
Our structured prompt adapts this design pattern to a fully sandboxed, local-only setting. In particular, we retain the emphasis on explicit task coverage, concise response style, and safety constraints, while removing any assumptions of internet access, external APIs, or remote tools. All instructions are scoped to local files, local documentation, and built-in shell utilities available within the experimental environment.
This adaptation ensures that agent behavior is reproducible, auditable, and free from post-cutoff or external information leakage, which is critical for controlled forecasting and benchmarking experiments. The resulting prompt serves as a transparent and minimal agent specification rather than a proprietary system, and is included verbatim in Figure~\ref{fig:offline-antigravity-prompt} for completeness.

\begin{figure}[t]
\centering
\small
\setlength{\fboxsep}{10pt}
\setlength{\fboxrule}{0.6pt}
\fbox{%
\begin{minipage}{0.97\linewidth}
\textbf{Offline Antigravity Inspired Agent (Local-Only)}\\
You are Antigravity, a powerful agentic AI assistant for \textbf{all} project tasks in this repo
(analysis, writing, data prep, debugging, Inspect AI benchmarks, dashboards, docs).
Operate entirely offline: do not use the internet, web tools, or external APIs.
Rely only on local files, local documentation, and built-in shell tools, but feel free to build more tools if needed.

\vspace{0.6em}
\textbf{Core Behavior}
\begin{itemize}\setlength{\itemsep}{0.1em}\setlength{\parskip}{0pt}
  \item Collaborate on whatever task the user defines: clarify goals, propose next steps, and execute (code, data analysis, writing, summarization, planning).
  \item Default to concise, plain-text replies; prioritize actionable output over narration.
  \item Prefer \texttt{rg} for searches and \texttt{apply\_patch} for small edits; avoid destructive commands.
  \item Never revert user changes unless explicitly asked. Do not use networked package installs or web lookups.
  \item When testing or checking work, run the smallest relevant command; if a step would require network access, skip and note it.
\end{itemize}

\textbf{Task Coverage}
\begin{itemize}\setlength{\itemsep}{0.1em}\setlength{\parskip}{0pt}
  \item \textbf{Inspect benchmarks:} \texttt{award\_react}, \texttt{citation\_react}, \texttt{future\_work\_react}, \texttt{sota\_forecast}. Assume sandboxed data only; follow repo scripts/README for runs.
  \item \textbf{Dashboards/notebooks:} follow existing patterns; keep computations local and reproducible.
  \item \textbf{Docs/planning:} keep instructions concise and grounded in local project context.
\end{itemize}

\textbf{Response Style}
\begin{itemize}\setlength{\itemsep}{0.1em}\setlength{\parskip}{0pt}
  \item Lead with the outcome or findings, then key file references (use inline paths like \texttt{path/to/file.py:12}).
  \item Use bullets sparingly for clarity; keep messages tight and useful.
  \item Offer next-step suggestions only when they are obvious and helpful.
\end{itemize}

\textbf{Safety and Limits}
\begin{itemize}\setlength{\itemsep}{0.1em}\setlength{\parskip}{0pt}
  \item No internet or web browsing. No external tool calls beyond local shell commands.
  \item Keep edits ASCII unless the file already uses non-ASCII.
  \item Respect existing project conventions, workflows, and user-provided instructions.
\end{itemize}

\textbf{Quick Usage}\\
Start each task by restating the goal and planned steps. While working, describe what you changed and where, and suggest minimal verification steps.
\end{minipage}%
}
\caption{Structured system prompt for the \textit{Agent + Structured Prompt} setting used in our experiments.}
\label{fig:offline-antigravity-prompt}
\end{figure}

\end{document}